\RequirePackage{fix-cm}
\documentclass[twocolumn]{svjour3}         
\smartqed  
\usepackage{savesym}
\savesymbol{widetilde}

\usepackage{pgfplots}

\usepackage{gensymb}

\usepackage{graphicx}
\usepackage{amsmath,amssymb} 

\restoresymbol{AMS}{widetilde}

\usepackage{widetext}

\usepackage{float}

\usepackage{url}
\usepackage{hyperref}
\usepackage{fancyhdr}

\def\makeheadbox{\relax}

\newcommand{\loss}{\mathcal{L}}
\newcommand{\img}{\vec{p}}

\newcommand{\frm}{\vec{f}}
\newcommand{\frmn}[1]{\frm^{(#1)}}
\newcommand{\gen}{\vec{x}}
\newcommand{\geni}{\vec{x'}}
\newcommand{\genn}[1]{\gen^{(#1)}}
\newcommand{\genni}[1]{\geni^{(#1)}}
\newcommand{\gennm}[2]{\genn{#1,#2}}
\newcommand{\gennmi}[2]{\genni{#1,#2}}

\newcommand{\styl}{\vec{a}}
\newcommand{\net}{\phi}
\newcommand{\model}{\Phi}
\newcommand{\modeli}{\Phi^{img}}
\newcommand{\modelv}{\Phi^{vid}}
\newcommand{\flow}{\vec{\omega}}
\newcommand{\cert}{\vec{c}}
\newcommand{\acert}{\overline{\cert}}
\newcommand{\certn}[2]{\cert^{(#1,#2)}}
\newcommand{\acertn}[2]{\acert^{(#1,#2)}}
\newcommand{\certi}{c}
\newcommand{\certlong}{\cert_{long}}
\newcommand{\certlongn}[2]{\certlong^{(#1,#2)}}
\newcommand{\warp}[2]{\omega_{#1}^{#2}}
\newcommand{\param}{\vec{w}}

\begin{document}
\sloppy

\title{Artistic style transfer for videos and spherical images\thanks{This study was partially
supported by the Excellence Initiative of the German Federal and State
Governments EXC 294}
}


\author{Manuel Ruder         \and
        Alexey Dosovitskiy \and
        Thomas Brox
}


\institute{M. Ruder, A. Dosovitskiy, T. Brox \at
           Department of Computer Science and BIOSS Centre for Biological Signalling Studies, University of Freiburg \\
              \email{\{rudera,dosovits,brox\}@cs.uni-freiburg.de}           
}


\maketitle

\begin{abstract}
Manually re-drawing an image in a certain artistic style takes a professional artist a long time.
Doing this for a video sequence single-handedly is beyond imagination.
We present two computational approaches that transfer the style from one image (for example, a painting) to a whole video sequence.
In our first approach, we adapt to videos the original image style transfer technique by Gatys et al. based on energy minimization. We introduce new ways of initialization and new loss functions to generate consistent and stable stylized video sequences even in cases with large motion and strong occlusion.
Our second approach formulates video stylization as a learning problem. We propose a deep network architecture and training procedures that allow us to stylize arbitrary-length videos in a consistent and stable way, and nearly in real time.
We show that the proposed methods clearly outperform simpler baselines both qualitatively and quantitatively.
Finally, we propose a way to adapt these approaches also to 360 degree images and videos as they emerge with recent virtual reality hardware. 
\keywords{Style transfer \and Deep networks \and Artistic videos \and Video stylization}
\end{abstract}

\section{Introduction}

\begin{figure*}[t]
	\centering
	\includegraphics[width=1\linewidth]{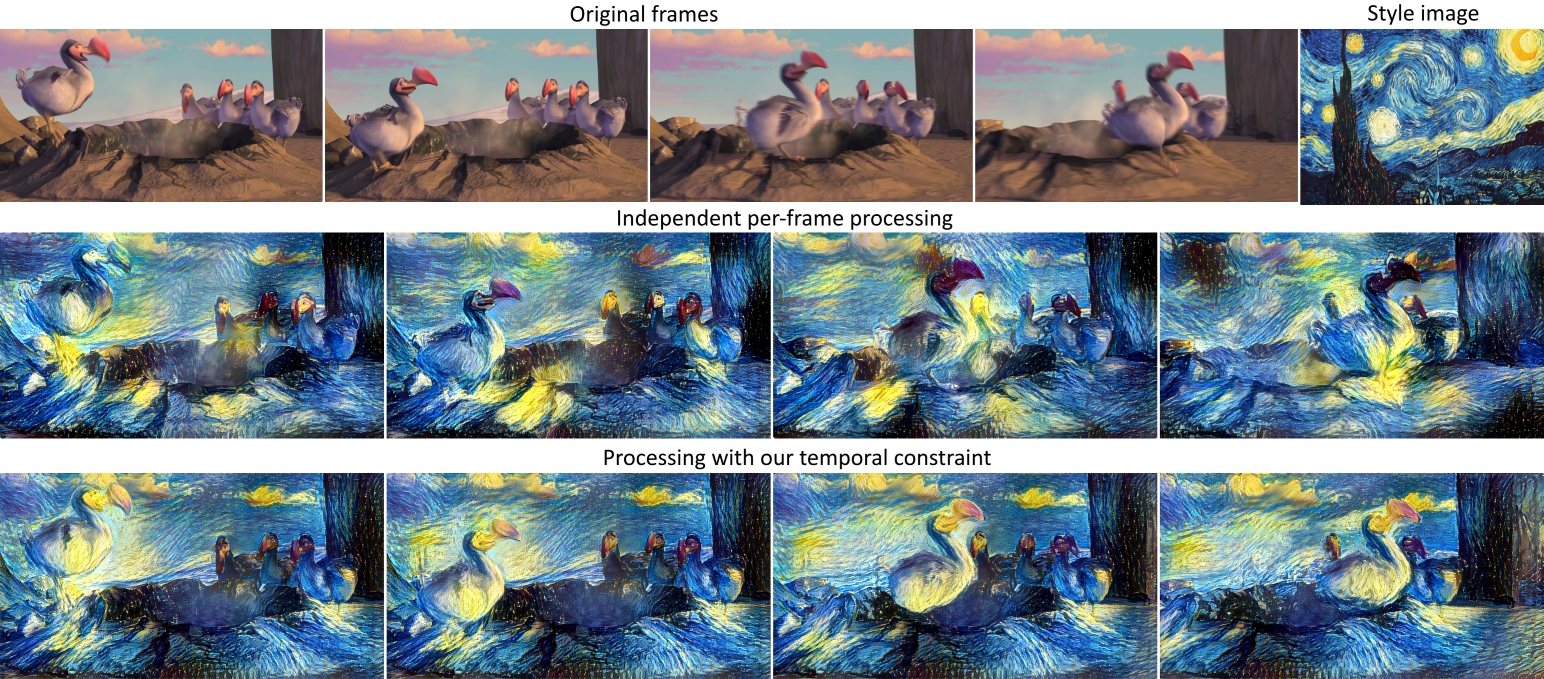}
	\caption{Scene from \emph{Ice Age} (2002) processed in the style of \emph{The Starry Night}. 
		Comparing independent per-frame processing to our temporally consistent approach, the latter is clearly preferable. Best observed in the supplemental video, see \url{https://youtu.be/SKql5wkWz8E}.}
	\label{fig:ImageTeaser}
\end{figure*}

The seminal work of Gatys et al.~\cite{Gatys2016} that transfers the artistic style of a painting to real world photographs has opened up an interesting application area for deep networks and has triggered lots of follow-up works. Their approach uses the statistics of high-level feature representations of the images from the hidden layers of an image classification network to separate and reassemble content and style. This is done by formulating an optimization problem that, starting with white noise, searches for a new image showing similar neural activations as the \textit{content image} and similar feature correlations (expressed by a Gram matrix) as the \textit{style image}. Based on this, Johnson et al.~\cite{Johnson2016} demonstrated that style transfer, conceived as an image transformation problem, can be learned by a convolutional neural network, which drastically reduces the runtime to process an image.

This paper presents extensions of the works from Gatys et al.~\cite{Gatys2016} and Johnson et al.~\cite{Johnson2016} to video sequences. Given an artistic image, we transfer its particular style of painting to the entire video. Processing each frame of the video independently leads to flickering and false discontinuities, since the solution of the style transfer task is not stable.
To regularize the transfer and to preserve smooth transition between individual frames of the video, we introduce a temporal constraint that penalizes deviations between two successive frames. The temporal constraint takes the optical flow from the original video into account: instead of penalizing the deviations from the previous frame, we penalize deviation along the point trajectories. Disoccluded regions as well as motion boundaries are excluded from the penalizer. This allows the process to rebuild disoccluded regions and distorted motion boundaries while preserving the appearance of the rest of the image; see Fig.~\ref{fig:ImageTeaser}.

Adding a loss that prefers temporal consistency between successive frames alone still leads to visible artifacts in the resulting videos. Thus, we present two extensions to our basic approach. The first one aims on improving the consistency over larger periods of time. When a region occluded in some frame and disoccluded later gets rebuilt during the process, most likely this region will be given a different appearance than before the occlusion. We solve this issue by taking long-term motion estimates into account. Secondly, the style transfer tends to create artifacts at the image boundaries. Therefore, we developed a multi-pass algorithm processing the video in alternating temporal directions using both forward and backward flow. 

Apart from the optimization-based approach, which yields high-quality results but is very slow to compute, we also present a network-based approach to video style transfer in the spirit of Johnson et al. \cite{Johnson2016}. We present a network to stylize subsequent frames of a video given the previously stylized frame as additional input, warped according to the optical flow. As with the optimization-based approach, we incorporate a temporal constraint into the loss and train a network which can be applied recursively to obtain a stable stylized video. 

We compare both approaches in terms of speed and consistency. We show quantitative results on the Sintel MPI dataset and qualitative results on several movie shots. We were able to successfully eliminate most of the temporal artifacts and can create smooth and coherent stylized videos both using the optimization-based approach and image transformation networks.
Additionally, we provide a detailed ablation study showing the importance of different components of both approaches.

Finally, we apply the style transfer to spherical images projected onto a cube map. The goal is to stylize a cube map such that adjacent edges remain consistent to each other. Technically this is related to video style transfer. We show that style transfer to videos and spherical images can be combined to process spherical videos. Although the approach is still a little too slow for real-time processing, it demonstrates the potential of artistic virtual reality applications.

\section{Related work}
\label{sec:introduction}
\textbf{Image style transfer.} In an initial study, Gatys et al. \cite{DBLP:conf/nips/GatysEB15} showed that hidden feature maps learned by a deep neural network can be used for texture synthesis by using feature correlations of an image classification network as a descriptor for the texture.

Later, Gatys et al. \cite{Gatys2016} extended their approach to synthesizing images which have the texture (or, more generally, the style) of one image while showing the content of another image.
The style is represented by feature correlations and the content of an image is described by high-level feature maps of the deep neural network.

The approach of Gatys et al. received much attention inside and outside the computer vision community and triggered a whole line of research on deep-learning-based style transfer. Li et al. \cite{DBLP:conf/cvpr/LiW16} proposed a different way to represent the style with the neural network to improve visual quality where the content and style image show the same semantic content. Nikulin et al. \cite{DBLP:journals/corr/NikulinN16} tried the style transfer algorithm on features from other networks and proposed several variations in the way the style of the image is represented to achieve different goals like illumination or season transfer. Luan et al. presented an approach for photo-realistic style transfer, where both the style image and the content are photographs \cite{luan2017deep}.

\textbf{Fast image style transfer.} The processing time can be reduced by orders of magnitude by training a deep neural network to learn a mapping from an image to its stylized version. Different approaches have been developed: Li et al. \cite{DBLP:conf/eccv/LiW16} used supervised training and adversarial training. Johnson et al. \cite{Johnson2016} presented an unsupervised training approach where the loss is computed by using neural network features and statistics as introduced by Gatys et al.~\cite{Gatys2016}. Ulyanov et al. \cite{DBLP:conf/icml/UlyanovLVL16} proposed an unsupervised approach, too, but used a multi-scale network architecture. 
Unsupervised approaches allow the use of larger training datasets, like Microsoft COCO, while Li et al. precomputed a small number of stylized images and generated more training samples by data augmentation techniques.
Ulyanov et al.~\cite{DBLP:journals/corr/UlyanovVL16} later introduced instance normalization, which greatly improves the qualitative results of feed-forward style transfer networks. Common to all approaches is the limitation that each model can only learn to transform the style according to one specific style template. However, recent techniques also allow multi-style networks \cite{DBLP:journals/corr/ZhangD17,Ghiasi2017}.

\textbf{Painted animations.} Litwinowicz \cite{Litwinowicz:1997:PIV:258734.258893} proposed an approach to create temporally consistent artistic videos by adding artificial brush strokes to the video. To gain consistency, brush strokes were generated for the first frame and then moved along the optical flow field. Different artistic styles were received by modifying various parameters, such as thickness, of these brush strokes, or by using different brush placement methods. Hays et al. \cite{Hays:2004:IVB:987657.987676} built upon this idea and proposed new stylistic parameters for the brush strokes to mimic different artistic styles. O'Donovan et al. \cite{odonovan:2012} introduced an energy minimization problem for this task. The optimization objective included a temporal constraint by penalizing changes in shape and width of the brush strokes compared to the previous frame.

\textbf{Video style transfer.} In an earlier conference version of the present paper, we demonstrated for the first time the capability of the optimization-based approach to stylize videos in a consistent way by incorporating additional loss functions into the objective function \cite{Ruder2016}. In the present paper, we extend this previous work to fast, network-based style transfer and to spherical images and videos.

Recently, Chen et al.  \cite{DBLP:journals/corr/ChenLYYH17}, too, implemented video style transfer using feed-forward style transfer networks. Different from this work, Chen et al. improved consistency by reusing features from the intermediate layers of the style transfer network from the previous frame: they are warped using optical flow and combined with feature maps extracted from the current frame in non-occluded regions before passed to the decoder part of the network.
More recently, Gupta et al.~\cite{Gupta2017} and Huang et al.~\cite{Huang17} presented style transfer networks that do not use optical flow at test time. Similar to the present work, temporal inconsistencies are penalized by a consistency loss based on the optical flow field. As input, however, their style transfer networks do not use the last stylized frame warped, but two video frames only (Huang et al.) or one frame in combination with the unwarped previous stylized frame (Gupta et al.).

\section{Style transfer in still images}\label{sec:style-transfer-still-images}
\label{sec:methods}

In this section, we briefly review the image style transfer approaches introduced by Gatys et al.~\cite{Gatys2016} and Johnson et al. \cite{Johnson2016}, which are used as basis for our video style transfer algorithms.

The goal is to generate a stylized image $\gen$ showing the content of an image $\img$ in the style of an image $\styl$.
Gatys et al. formulated an energy minimization problem consisting of a \textit{content loss} and a \textit{style loss}.
The key idea is that features extracted by a convolutional network carry information about the content of the image, while the correlations of these features encode the style. 

We denote by $\net^l (\cdot)$ the function implemented by the part of the convolutional network from input up to layer $l$.
We denote the feature maps extracted by the network from the original image $\img$, the style image $\styl$, and the stylized image $\gen$ by $\vec{P}^l = \net^l (\img)$, $\vec{S}^l = \net^l(\styl)$ and $\vec{F}^l = \net^l(\gen)$, respectively.
The dimensionality of these feature maps we denote by $N_l \times M_l$, where $N_l$ is the number of filters (channels) in the layer, and $M_l$ is the spatial dimensionality of the feature map, that is, the product of its width and height.

The content loss, denoted as $\loss_{content}$, is the mean squared error between 
$\vec{P}^l \in \mathbb{R}^{N_l\times M_l}$ 
and $\vec{F}^l \in \mathbb{R}^{N_l\times M_l}$. 
This loss need not be restricted to only one layer. Let $L_{content}$  be the set of layers to be used for content representation, then we have:
\begin{equation}
\loss_{content}\bigl(\vec{p}, \vec{x}\bigr) = \sum_{l \in L_{content}}  \frac{1}{N_l M_l} \sum_{i,j} \bigr( F^l_{ij} - P^l_{ij} \bigl)^2.
\end{equation}

The style loss is also a mean squared error, but between the correlations of the filter responses expressed by their Gram matrices $A^l \in \mathbb{R}^{N_l\times N_l}$ for the style image $\styl$ and $G^l \in \mathbb{R}^{N_l\times N_l}$ for the stylized image $\gen$.
These are computed as
$A^l_{ij} = \sum\limits_{k=1}^{M_l} S^l_{ik} S^l_{jk}$ and 
$G^l_{ij} = \sum\limits_{k=1}^{M_l} F^l_{ik} F^l_{jk}\,$.
As above, let  $L_{style}$ be the set of layers we use to represent the style, then the style loss is given by:
\begin{equation}
\mathcal{L}_{style}\bigl(\vec{a}, \vec{x}\bigr) = \sum_{l \in L_{style}} \frac{1}{N_l^2 M_l^2} \sum_{i,j} \bigr( G^l_{ij} - A^l_{ij} \bigl)^2 
\end{equation}
Overall, the loss function is given by
\begin{equation}
\mathcal{L}_{image}\bigl(\vec{p}, \vec{a}, \vec{x}\bigr) =
\alpha \mathcal{L}_{content}\bigl(\vec{p}, \vec{x}\bigr) \ + \
\beta \mathcal{L}_{style}\bigl(\vec{a}, \vec{x}\bigr),
\end{equation}
with weighting factors $\alpha$ and $\beta$ governing the relative importance of the two components.

The stylized image is computed by minimizing this energy with respect to $\gen$ using gradient-based optimization.
Typically it is initialized with random Gaussian noise.
However, the loss function is non-convex, therefore the optimization is prone to falling into local minima. 
This makes the initialization of the stylized image important, especially when applying the method to frames of a video.\\

Instead of solving an optimization problem, a much faster method directly learns a style transfer function for one particular style mapping from an input image to its stylized correspondence. Such a function $\model_{\param}$ can be expressed by a convolutional neural network with parameters $\param$. The network is trained to minimize the expected loss for an arbitrary image~$\img$:

\begin{equation}
\param^* = \arg\min_{\param} \textbf{E}_{\img} \Biggl[ \loss_{image} \Bigl( \img, \styl, \model_{\param}(\img) \Bigr) \Biggr].
\end{equation}

Johnson et al. \cite{Johnson2016} directly evaluate $\loss_{image}$ in each iteration of the training phase and use the loss and gradients from this function to perform a backward pass. Since this function does not compare the output of the network to a ground truth but acts as a perceptual quality measure, this is called a perceptual loss function.\\
The network architecture chosen by Johnson et al. \cite{Johnson2016} is a fully convolutional neural network, which uses downsampling by strided convolution to increase the receptive field and upsampling to produce the final output. Furthermore, the network utilizes residual connections which help the network converge faster.\\

The quality of network-based style transfer can be significantly improved with instance normalization~\cite{DBLP:journals/corr/UlyanovVL16}~-- a technique inspired by batch normalization~\cite{Ioffe2015BatchNA}. Instance normalization performs contrast normalization for each data sample. Details are provided in the Appendix.

\section{Optimization-based coherent video style transfer}\label{sec:slow-video-transfer}

In this section, we explain our video style transfer approach formulated as an energy minimization problem. This includes two extensions that improve long-term consistency and image quality during camera motion.

We use the following notation: $\frmn{i}$ is the $i^{th}$ frame of the original video, $\vec{a}$ is the style image and $\genn{i}$ are the stylized frames to be generated. Furthermore, we denote by $\genni{i}$ the initialization of the style optimization algorithm at frame $i$. 
By $x_j$ we denote the $j^{th}$ component of a vector $\gen$.

\subsection{Short-term consistency by initialization}

Independent initialization with Gaussian noise yields two consecutive frames that are stylized very differently, even when the original frames hardly differ.
The most basic way to improve temporal consistency is to initialize the optimization for the frame $i+1$ with the already stylized frame $i$. Areas that have not changed between the two frames are then initialized with the desired appearance, while the rest of the image has to be rebuilt through the optimization process.

This simple approach is insufficient for moving parts of the scene, because the initialization does not match.
Thus, we take the optical flow into account and initialize the optimization for frame $i+1$ with the previous stylized frame warped by the optical flow: $\genni{i+1} = \warp{i}{i+1}\bigl(\genn{i}\bigr)$. 
Here $\warp{i}{i+1}$ denotes the function that warps a given image using the optical flow field that was estimated between images $\frmn{i}$ and $\frmn{i+1}$.
Only the first frame of the stylized video is initialized randomly.

Every reliable optical flow technique can be used for computing the optical flow. We experimented with two popular algorithms: DeepFlow \cite{weinzaepfel:hal-00873592} and EpicFlow \cite{revaud:hal-01142656}.
Both are based on Deep Matching~\cite{weinzaepfel:hal-00873592}: DeepFlow combines it with a variational approach, while EpicFlow relies on edge-preserving sparse-to-dense interpolation. 

\subsection{Temporal consistency loss} \label{sec:video-loss}
To enforce stronger consistency between adjacent frames we additionally introduce an explicit consistency penalty to the objective function. Disoccluded regions as well as motion boundaries are excluded from the penalizer, which allows the optimizer to rebuild those regions.
To detect disocclusions, we perform a forward-backward consistency check of the optical flow \cite{Bro10e}.
Let $\flow = (u, v)$ be the optical flow in forward direction and $\hat{\flow} =(\hat{u}, \hat{v})$ the flow in backward direction.
Denote by $\widetilde{\flow}$ the forward flow warped to the second image:
\begin{equation}
\widetilde{\flow} (x,y) = \flow((x,y) + \hat{\flow}(x,y)).
\end{equation}

In areas without disocclusion, this warped flow should be approximately the opposite of the backward flow. Therefore we mark as disocclusions those areas where the following inequality holds:
\begin{equation} \label{eq:disoccl}
|\widetilde{\flow}+ \hat{\flow}|^2 > 0.01 (|\widetilde{\flow}|^2 + |\hat{\flow}|^2) + 0.5
\end{equation}
Motion boundaries are detected using the following inequality:
\begin{equation} \label{eq:motionb}
|\nabla \vec{\hat{u}}|^2 + |\nabla \vec{\hat{v}}|^2 > 0.01 |\vec{\mathrm{\hat{w}}}|^2  + 0.002
\end{equation}
Coefficients in inequalities~\eqref{eq:disoccl} and~\eqref{eq:motionb} are taken from Sundaram et al.~\cite{Bro10e}.

The temporal consistency loss function penalizes deviations from the warped image in regions where the optical flow is consistent and estimated with high confidence:
\begin{equation}
\mathcal{L}_{temporal}(\vec{x}, \vec{\omega}, \cert) = \frac{1}{D} \sum_{k=1}^D \certi_k \cdot (x_k - \omega_k)^2\;.
\end{equation}
Here $\cert \in [0, 1]^D$ is a per-pixel weighting of the loss and $D = W \times H \times C$ is the dimensionality of the image. We define the weights $\cert^{(i-1,i)}$ between frames $i\!-\!1$ and $i$ as follows: $0$ in disoccluded regions (as detected by forward-backward consistency) and at the motion boundaries, and $1$ elsewhere. 
The overall loss takes the form:
\begin{align}
&\loss_{shortterm}\bigl(\frmn{i}, \styl, \genn{i}\bigr) =& \\
&\ \ \alpha \mathcal{L}_{content}\bigl(\frmn{i}, \genn{i} \bigr) \ + \
\beta \loss_{style}\bigl(\styl, \genn{i}\bigr) \ + \notag\\
&\ \ \gamma \loss_{temporal}\bigl(\genn{i}, \warp{i-1}{i}(\genn{i-1}), \certn{i-1}{i}\bigr)\;.
\end{align}

We optimize the frames sequentially, thus $\genn{i-1}$ refers to the already stylized frame $i\!-\!1$.

We also experimented with the more robust absolute error instead of squared error for the temporal consistency loss; results are shown in section~\ref{sec:robust_loss}.

\subsection{Long-term consistency}
The short-term model only accounts for consistency between adjacent frames. Therefore, areas that are occluded in some frame and disoccluded later will likely change their appearance in the stylized video. This can be counteracted by taking the long-term motion into account, i.e., not only penalizing deviations from the previous frame, but also from temporally distant frames. Let $J$ denote the set of indices each frame should take into account relative to the frame number, e.g., with $J = \{1, 2, 4\}$, frame $i$ takes frames $i\!-\!1$, $i\!-\!2$, and $i\!-\!4$ into account. The loss function with long-term consistency is given by:
\begin{align}
&\mathcal{L}_{longterm}\bigl(\frmn{i}, \styl, \genn{i}\bigr) = \ \notag\\
&\ \ \alpha \mathcal{L}_{content}\bigl(\frmn{i}, \genn{i}\bigr) \
+ \ \beta \mathcal{L}_{style}\bigl(\styl, \genn{i}\bigr) \ + \notag\\
& \ \ \gamma \sum_{j\in J: i-j \ge 1} \mathcal{L}_{temporal}\bigl(\genn{i},  \warp{i-j}{i}(\genn{i-j}), \certlongn{i-j}{i}\bigr)
\end{align}
The weights $\certlongn{i-j}{i}$ are adjusted such that each pixel is only connected to the closest possible frame from the past. Since the optical flow computed over more frames is more erroneous than over fewer frames, this results in nicer videos. This is implemented as follows: let $\certn{i-j}{i}$ be the weights for the flow between image $i\!-\!j$ and $i$, as defined for the short-term model. The long-term weights $\certlongn{i-j}{i}$ are then computed as:
\begin{equation}
\certlongn{i-j}{i} = \max \bigl( \certn{i-j}{i} - \sum_{k\in J: i-k > i-j} \certn{i-k}{i},\; \vec{0} \bigr) \; ,
\end{equation}
where the maximum is applied element-wise.
This means, we first apply the usual short-term constraint. For pixels in disoccluded regions we look into the past until we find a frame in which these have consistent correspondences.
An empirical comparison of $\cert^{(i-j,i)}$ and $\certlong^{(i-j,i)}$ is shown in the supplementary video (see section~\ref{sec:video}).

\subsection{Multi-pass algorithm}

We found that the output image tends to have less contrast and to be less diverse near image boundaries than in other areas of the image. This effect appears particularly in dynamic videos with much camera motion. Disocclusion at image boundaries constantly creates small areas with new content that moves towards other parts of the image. This leads to a lower image quality over time when combined with our temporal constraint. Therefore, we developed a multi-pass algorithm that processes the whole sequence in multiple passes and alternates between the forward and backward direction. Every pass consists of a relatively low number of iterations without full convergence. At the beginning, we process every frame independently. After that, we blend frames with non-disoccluded parts of previous frames warped according to the optical flow, then run the optimization algorithm for some iterations initialized with this blend.
We repeat this blending and optimization to convergence.

Let $\gennmi{i}{j}$ be the initialization of frame $i$ in pass $j$ and $\gennm{i}{j}$ the corresponding output after some iterations of the optimization algorithm.
When processed in forward direction, the initialization of frame $i$ is created as follows:
\begin{equation}
\gennmi{i}{j} = 
\begin{cases}
\gennm{i}{j-1} &\text{if } i=1,\\
\delta \certn{i-1}{i} \circ \warp{i-1}{i}\bigl(\gennm{i-1}{j}\bigr)+  \\
(\overline{\delta} \vec{1} + \delta \acertn{i-1}{i}) \circ \gennm{i}{j-1} &\text{else}.
\end{cases}
\end{equation}
Here $\circ$ denotes element-wise vector multiplication, $\delta$ and $\overline{\delta} = 1\!-\!\delta$ are the blend factors, $\vec{1}$ is a vector of all ones, and $\overline{\cert} = \vec{1} - \cert$.

Analogously, the initialization for a backward direction pass is:
\begin{equation}
\gennmi{i}{j} = 
\begin{cases}
\gennm{i}{j-1} &\text{if } i=N_{\text{frames}}\\
\delta \certn{i+1}{i} \circ \warp{i+1}{i}\bigl(\gennm{i+1}{j}\bigr)+  \\
(\overline{\delta} \vec{1} + \delta \acertn{i+1}{i}) \circ \gennm{i}{j-1}
&\text{else.}
\end{cases}
\end{equation}

The multi-pass algorithm can be combined with the temporal consistency loss described above. We achieved good results when we disabled the temporal consistency loss in several initial passes and enabled it in later passes after the images had stabilized.

\begin{figure}
	\centering
	\includegraphics[width=1.0\linewidth]{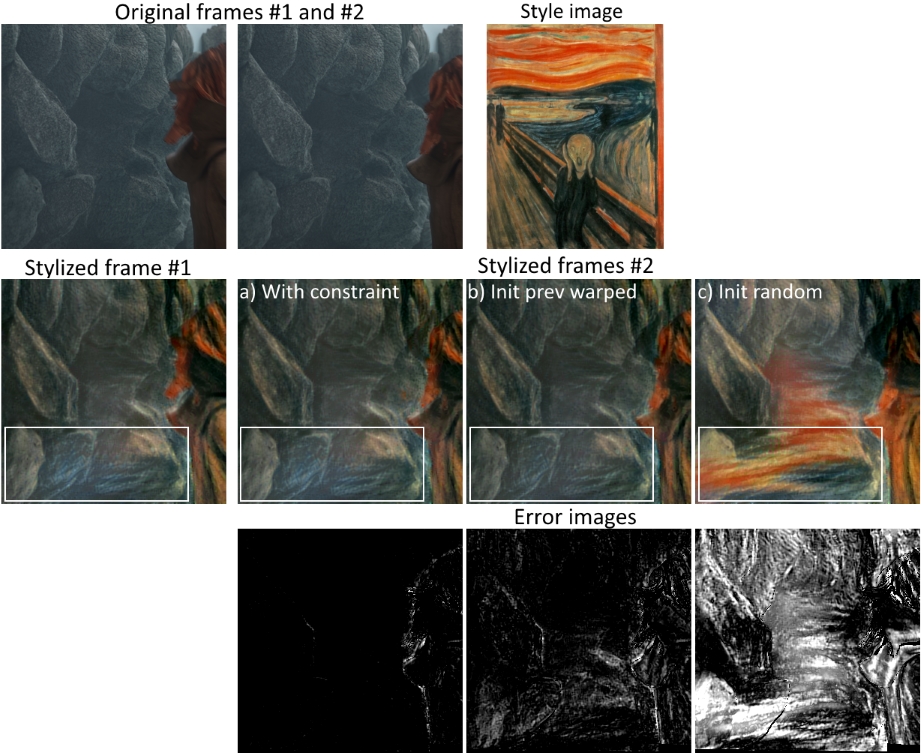}
	\caption{Close-up of a scene from Sintel, combined with \emph{The Scream} painting. \textbf{a)} With temporal constraint \textbf{b)} Initialized with previous image warped but without the constraint \textbf{c)} Initialized randomly. The marked regions show most visible differences. \emph{Error images} show the contrast-enhanced absolute difference between frame \#1 and frame \#2 warped back using ground truth optical flow, as used in our evaluation. The effect of the temporal constraint is clearly visible in the error images.}
	\label{fig:sintel_qualitative}
\end{figure}

\section{Fast coherent video style transfer}\label{sec:fast-video-transfer}

\begin{figure*}
	\centering\includegraphics[width=0.825\textwidth]{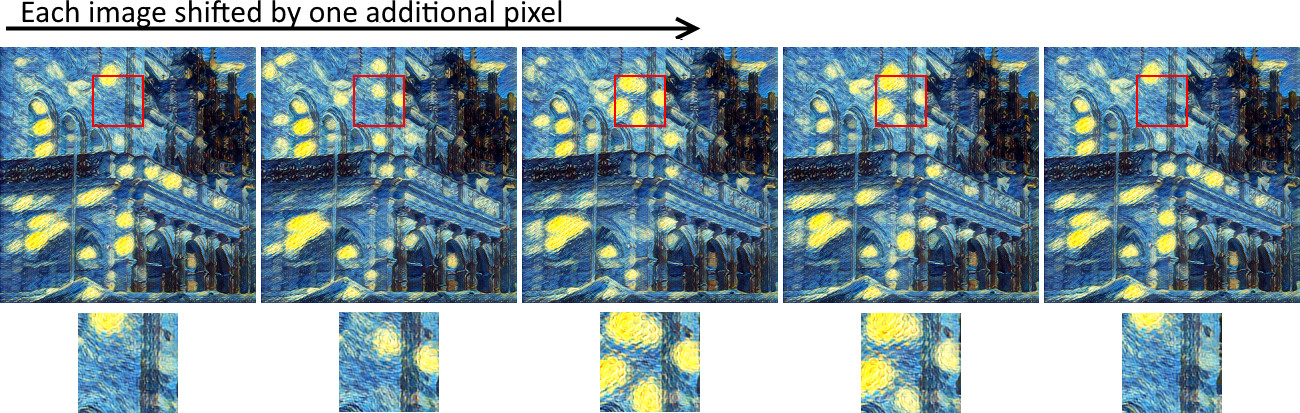}
	\caption[Analysis of flickering]{Images generated by feed-forward style transfer. Starting from the left, each image is shifted by one additional pixel in x and y direction. The first and the fifth image, where the shift accumulates to a multiple of four pixels, are stylized nearly identically. The other images are stylized very differently from the first image. This shows that there is no shift invariance unless the shift exactly matches the stride used within the network.}
	\label{fig:pixel-shift}
\end{figure*}

\subsection{Consistency by architecture}
Before introducing temporal consistency constraints into a network based approach, we investigated whether the temporal consistency of the output of the stylization network can be improved simply by using an architecture that is robust to transformations of the input image.
Figure~\ref{fig:pixel-shift} shows that the network of Johnson et al. is invariant to shifts of exactly four pixels, but its output is significantly disturbed by smaller shifts. 
The loss of shift invariance is caused by the use of downsampling (strided) layers in the network, with a total downsampling factor of $4$.

Shift invariance can be retained by avoiding downsampling and using a different technique for multi-scale context aggregation.
This can be achieved with dilated convolutions~\cite{YuKoltun2016}, which expand the convolutional filters instead of downsampling the feature maps. 
The experiments in Section~\ref{sec:Dilated} show that this indeed leads to improved temporal consistency, but at the cost of a heavily increased computational burden, especially in terms of memory.
This makes the approach impractical under most circumstances. 
Furthermore, this technique does not account for transformations that are more complex than axis-aligned shifts. 
Thus, in the following we present more efficient and more general techniques to improve temporal consistency in network based stylization.

\subsection{Training with the prior image}

Our main approach builds upon the architecture by Johnson et al. \cite{Johnson2016} but extends the network to take additional inputs. In addition to the content image, the network gets an incomplete image representing prior knowledge about the output, which we refer to as \textit{prior image}, as well as a per-pixel mask telling the network which regions of the prior image are valid. For video stylization, the prior image is the previously stylized frame warped according to the optical flow we calculated for the original video sequence. The mask is given by the occlusions between two frames and the motion boundaries. Occlusions are detected through a forward-backward consistency check, as described in the previous section, i.e., we calculate the optical flow both in forward and backward direction and test whether both estimates agree. At pixels where the estimates do not agree, and at motion boundaries, we set the certainty mask to zero. Everywhere else, we set this mask to $1$. \\

The goal of the network is to produce a new stylized image, where the combination of perceptual loss and deviation from the prior image (in non-occluded regions) is minimized.\\
The network $\modelv_{\param}$ with parameters $\param$ takes as input a frame $\frmn{t}$, the previously generated frame warped and masked by the occlusion mask $\warp{t-1}{t}(\genn{t-1})$ and yields the output $\genn{t}$. We can apply this function recursively to obtain a video. The first frame in the sequence is generated by an image network.

\begin{equation}
\genn{t} =
\begin{cases}
\modeli\left(\frmn{t} \right) & \quad \text{if } t = 1\\
\modelv\left(\frmn{t}, \ \warp{t-1}{t}(\genn{t-1}), \ \certn{t-1}{t}\right) & \quad \text{if } t > 1\\
\end{cases}
\end{equation}

Ideally, network parameters $\param$ would minimize the video loss function, as defined in Section \ref{sec:video-loss}, for an arbitrary video with $N_{frames}$ frames:

\begin{align}
\param^* = \arg\min_{\param} \textbf{E}_{\{\frmn{t}\}} \Biggl[  \sum_{t=2}^{N_{\text{frames}}} \loss_{video} \Bigl( \frmn{t}, \, \styl, \, \genn{t}, \, \genn{t-1} \Bigr) \Biggr], \notag\\
\text{with }\genn{t>1} = \modelv_{\param}\Bigl(\frmn{t}, \warp{t-1}{t}(\genn{t-1}), \certn{t-1}{t}\Bigr),
\end{align}
where $\genn{1} = \modeli_{\param'}(\frmn{1})$ for a given pre-trained parameter set $\param'$.

\begin{figure}
	\includegraphics[width=1.03\linewidth]{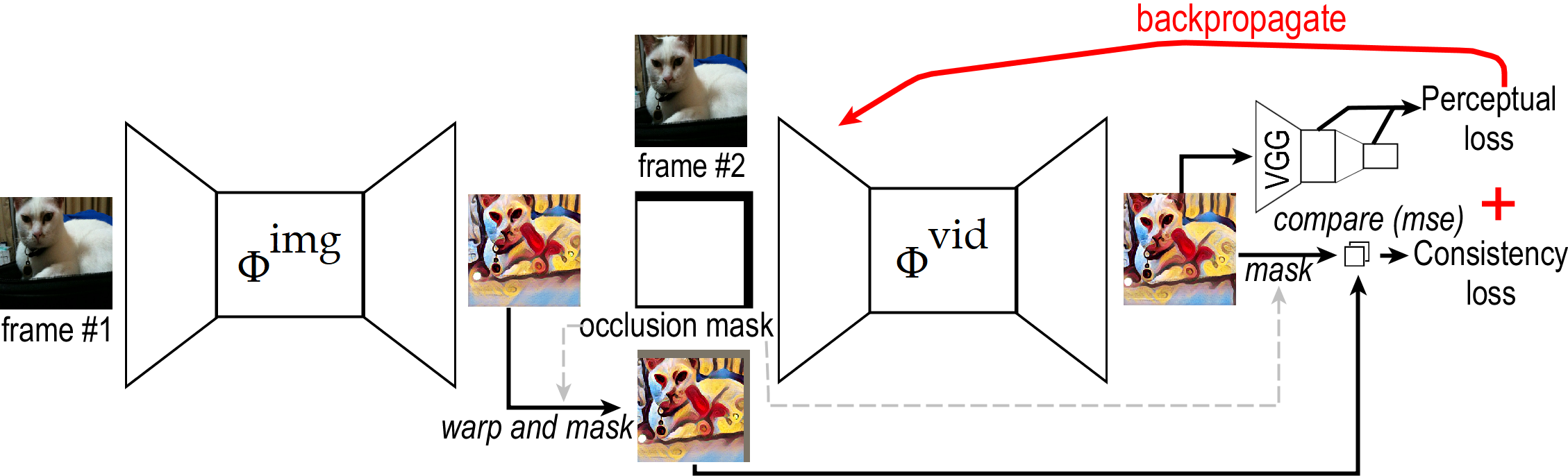}
	\caption[Training procedure two-frame.]{Basic training procedure for a style transfer network with prior image (e.g. last frame warped).}
	\label{fig:training-process}
\end{figure}

However, joint optimization for a whole video is computationally infeasible. 
Hence, we resort to recursively optimizing two-frame losses.
Our basic approach is illustrated in Figure~\ref{fig:training-process}. The network is trained to generate a successor frame to the first frame of a video:

\begin{align} \label{eq:twoframe}
\param^* = \arg\min_{\param} \textbf{E}_{\frmn{1},\frmn{2}} \Biggl[ \loss_{video} \Bigl( \frmn{2}, \ \styl, \ \genn{2}, \ \genn{1} \Bigr) \Biggr], \notag\\
\text{with }\genn{2} = \modelv_{\param}\Bigl(\frmn{2}, \warp{1}{2}(\genn{1}), \certn{1}{2}\Bigr).
\end{align}

We found this approach to yield unsatisfactory results when applied recursively to stylize several successive frames of a video. 
While the network learns to fix occlusions and to preserve the prior image as much as possible, there is nothing that prevents the propagation of errors and the degeneration of the image quality over time. In particular, we observed blurring and the loss of content accuracy in the image. In the following we introduce two improvements to avoid the degeneration of image quality.

\begin{figure*}
	\includegraphics[width=0.99\textwidth]{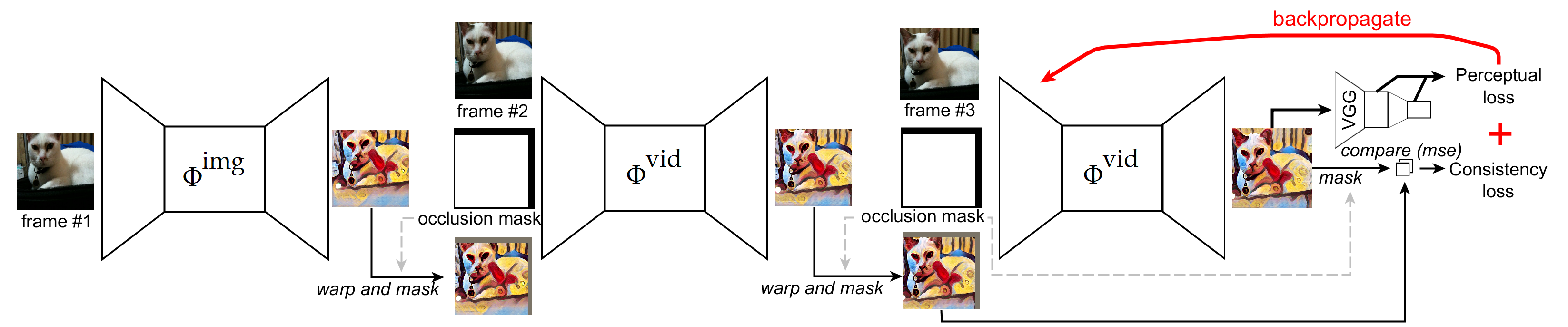}
	\caption[Training procedure multi-frame.]{Training procedure for our multi-frame approach, here shown with three frames. We found that back-propagating only one frame already improves the quality of generated videos a lot. Back-propagating more frames would have required decreasing the size of the network due to memory restrictions.}
	\label{fig:training-process-mf}
\end{figure*}

\subsubsection{Mixed training} \label{sec:mixed-training}

Our first approach to counteract the propagation of errors aims at forcing the network to refine the stylization at every step, rather than just copying the prior image. To this end, we randomly add training samples without prior image, i.e., the network has to generate a completely new image. We set the occlusion mask to zero everywhere, which also disables the temporal loss. Formally, the objective function changes to

\begin{align} \label{eq:mixed}
\param^* = \arg\min_{\param} \textbf{E}_{\frmn{1},\frmn{2}} \Biggl[ \loss_{video} \Bigl( \frmn{2}, \ \styl, \ \genn{2}, \ \genn{1} \Bigr) \Biggr] \notag\\
+ \textbf{E}_{\img} \Biggl[ \loss_{video} \Bigl( \img, \ \styl, \ \modelv_{\param}\Bigl(\img, \vec{0}, \vec{0}\Bigr), \ \vec{0} \Bigr) \Biggr] \notag\\
\text{with }\genn{2}=\modelv_{\param}\Bigl(\frmn{2}, \warp{1}{2}(\genn{1}), \certn{1}{2}\Bigr)
\end{align}
where $\vec{0}$ is a vector of the same dimension as the input image with zeros everywhere.

We increase the weight for the consistency loss proportionally to the percentage of empty-prior images. This aims for still achieving temporal consistency at test time, while focusing on content generation during training. 
We experimented with various fractions of zero-prior samples, and got good results when we set this fraction to $0.5$.

\subsubsection{Multi-frame training}

Our second approach to improve longer-term stylizations presents more than just the first two frames of a sequence during training. We rather recursively pass $\frmn{1}$ to $\frmn{n-1}$ through the network, i.e., training includes the potentially degenerated and blurred output, $\genn{n-1}$, as the prior image and requires the network to generate the non-degenerated ground truth $\genn{n}$. This forces the network to learn not only to copy the prior image but also to correct degeneration. If the network did not correct errors and blurriness in the prior image, this would lead to a large loss. Thus, the overall loss function eventually overrules the temporal consistency constraint. 

Formally, we generalize the objective function shown in Equation~\eqref{eq:twoframe} to

\begin{align} \label{eq:mf}
\param^* = \arg\min_{\param} \textbf{E}_{\frmn{t},\frmn{t+1}} \Biggl[ \loss_{video} \Bigl( \frmn{t+1}, \ \styl, \ \genn{t+1}, \ \genn{t} \Bigr) \Biggr], \notag\\
\text{with }\genn{t+1} = \modelv_{\param}\Bigl(\frmn{t+1}, \warp{t}{t+1}(\genn{t}), \certn{t}{t+1}\Bigr).
\end{align}

Since the network does not produce meaningful stylizations in the beginning of the training phase, we first start training the network with $t = 1$ and gradually increase $t$ in later iterations.

\subsubsection{Training data}

We used the Hollywood2 video scene dataset \cite{DBLP:conf/cvpr/MarszalekLS09}, a collection of 570 training videos. Some videos contain multiple shots. We split videos along shots and extracted $4,\!193$ distinct shots from that dataset. Since the dataset contains many frame pairs with little motion, we only took the five most differing frame tuples per shot. In more detail, we computed the squared distance between subsequent frames per tuple and took those five tuples which ranked highest on that measure.
In total, we acquired $20,\!965$ frame tuples. The size of a tuple corresponds to the five steps in our multi-frame training approach.

Additionally, we created a large synthetic set of video sequences based on the Microsoft COCO dataset \cite{coco04} with $80,\!000$ training images to increase the diversity in our training data.
To create each of these sequences, we selected a window from the image and moved it between $0$ to $32$ pixels in either direction to create consecutive frames.
Moreover, we used a zoom out effect, i.e., the window size was increased by up to $32$ pixels in either direction when generating a following frame, with all crops being resized to the dimensions of the original window size.
Translation strength and resize factor were sampled uniformly and independently for each axis for each training sample at training time.
The advantage of the synthetic data is that it comes with ground truth optical flow and occlusion maps.

\section{Spherical images and videos}

\begin{figure}[ht!]
	\centering
	\includegraphics[width=1.0\linewidth]{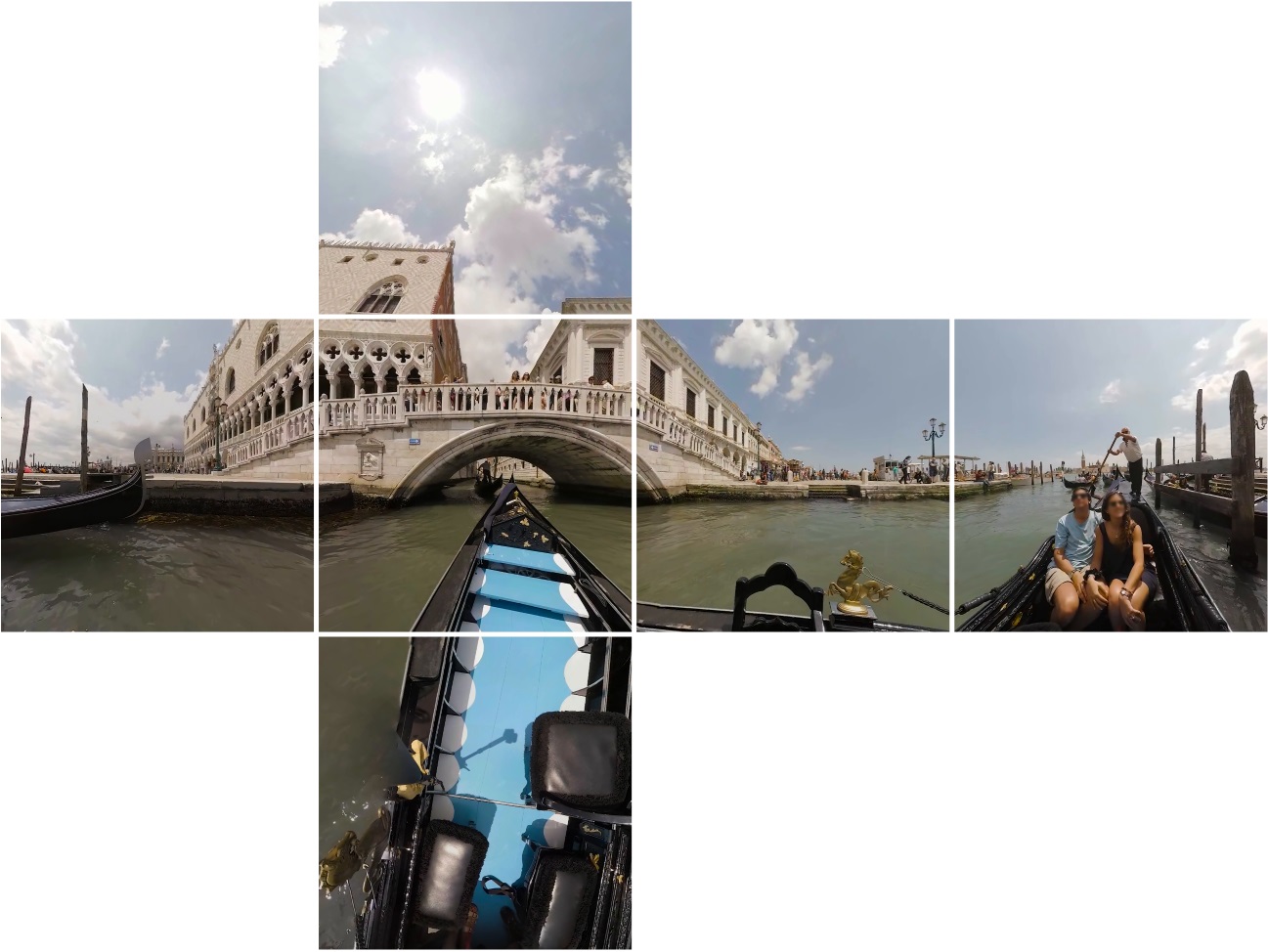}
	\caption[Cubemap projection]{Cubemap projection used for stylizing spherical images in this paper. The generated images must be consistent along the boundaries of neighboring cube faces. Every cube face has four neighbors.}
	\label{fig:Cubemap}
\end{figure}

Virtual reality (VR) applications become increasingly popular, and the demand for image processing methods applicable to spherical images and videos rises. Spherical reality media is typically distributed via a 2D projection. The most common format is the equirectangular projection. However, this format does not preserve shapes: the distortion of the projection becomes very large towards the poles of the sphere. Such non-uniform distortions are problematic for style transfer. Therefore, we work with subdivided spherical images that consist of  multiple rectilinear projections. In particular, we will use cubic projection, which represents a spherical image with six non-distorted square images, as illustrated in Figure~\ref{fig:Cubemap}. 
For style transfer in this regime, the six cube faces must be stylized such that their cut edges are consistent, i.e., the style transfer must not introduce false discontinuities along the edges of the cube in the final projection. Since applications in VR environments must run in real time, we only consider the fast, network-based approach in this section.  

\subsection{Border consistency}

\begin{figure}[!ht]
	\centering
	\includegraphics[width=0.7\linewidth]{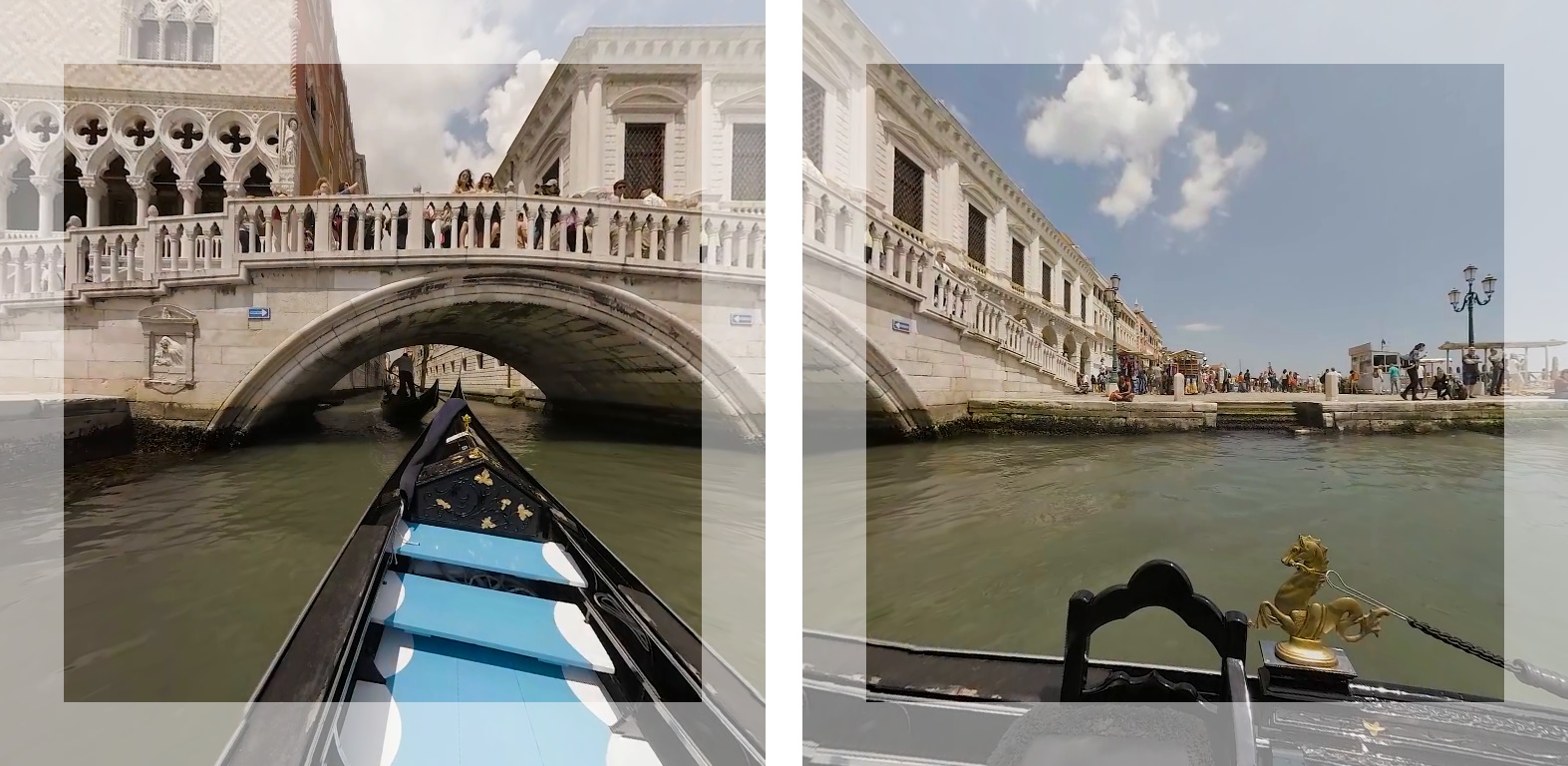}\\ \vspace{3mm}
	\includegraphics[width=1.0\linewidth]{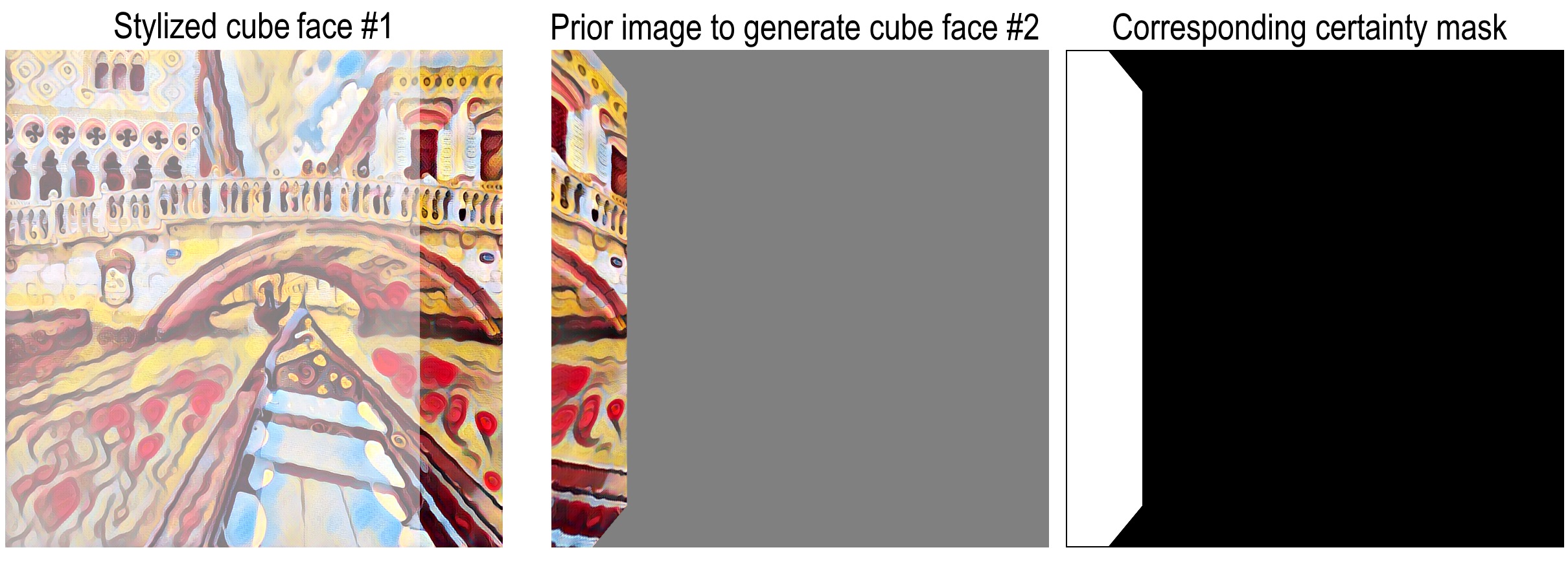}
	\caption{\textbf{Top:} Two extended cube faces of a spherical video with overlapping border regions, \textbf{bottom:} A stylized cubeface and the prior image generated from that.}
	\label{fig:VRData}
\end{figure}

In order to implement the consistency constraint, we extend the cube faces such that they overlap each other in the proximity of the face boundaries; see Figure \ref{fig:VRData} (top).\\

We stylize the cube faces sequentially and enforce subsequent cube faces to be consistent with already stylized edges from adjacent cube faces. This is similar to the video style transfer network, and we can use the same techniques described in detail in the previous section, such as mixed training and multi-frame training. Instead of using a previously generated frame from a video as the prior image, we use regions of adjacent cube faces overlapping with the current cube faces as prior image. The regions are transformed to match the cube face projection; see Figure \ref{fig:VRData} (bottom) and the Appendix.

\subsection{Training dataset} \label{sec:vr-training-data}

A style transfer network trained with regular video frames did not produce good results for stylizing cube faces, where a perspective transformed border of the image is given as the prior image. Therefore, we extended our dataset by training samples particularly designed for this task.

\begin{figure}[!ht]
	\centering
	\includegraphics[width=1.0\linewidth]{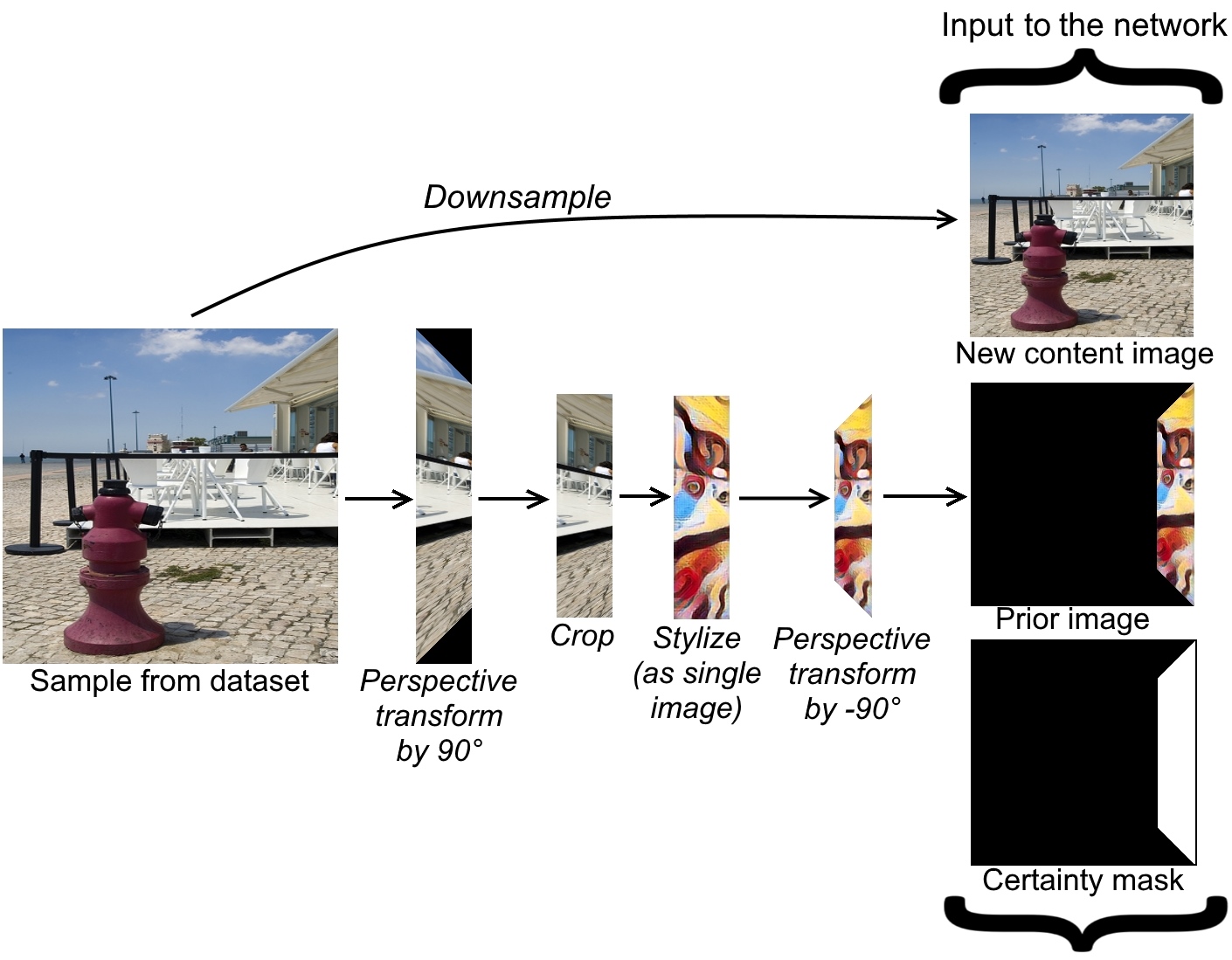}
	\caption[Spherical image training data.]{Illustration of the training data generation process for a network to adapt to perspectively transformed border regions.}
	\label{fig:vr-training}
\end{figure}

We stylized and transformed the image borders of samples from the Microsoft COCO dataset in such a way that it mimics the perspective transformed borders of a cube face prior image. This process is illustrated in Figure \ref{fig:vr-training}. We created a training set with $320,000$ samples (four samples for each image in COCO) this way. 

\subsection{Spherical videos}

The extensions for video style transfer and for spherical images can be combined to process spherical videos. 
This yields two constraints: (1) each cube face should be consistent along the motion trajectory; (2) neighboring cube faces must have consistent boundaries.

To implement the first constraint, we calculate optical flow for each cube face separately, which allows us to warp stylized cube faces the same way as for regular planar videos. To implement the second constraint, we blend both the warped image from the last frame and the transformed border of already stylized neighboring cube faces.

\section{Experiments on video style transfer}\label{chap:experiments}

In this section, we present the experimental results obtained with different versions of the presented methodology. Since it is hard to visualize qualitative differences in video content and spherical images appropriately on paper, we additionally refer to the supplemental video linked in Section~\ref{sec:video}.

\subsection{Dataset}

We used the MPI Sintel Dataset~\cite{Butler:ECCV:2012} including 23 training scenes with 20 to 50 frames of resolution $1024 \times 436$ pixels per scene. To keep the runtime at an acceptable level, we tested our optimization-based approach with a subset of five diverse scenes. Our feed-forward approach, being orders of magnitudes faster, was additionally evaluated on the full set.

The Sintel dataset provides ground truth optical flow and ground truth occlusion areas, which allows a quantitative study. To evaluate the short-term consistency, we warped each stylized frame at time $t$ back with the ground truth flow and computed the mean squared difference to the stylized frame $t-1$ in regions without disocclusion.
Each method was evaluated with different style templates, as shown in Section~\ref{sec:styles}.

\subsection{Implementation details}

\subsubsection{Optimization-based approach} \label{sec:details-optimization-approach}

Our implementation\footnote{\url{https://github.com/manuelruder/artistic-videos}} is based on a Torch~\cite{Collobert_NIPSWORKSHOP_2011} implementation called \emph{neural-style}\footnote{\url{https://github.com/jcjohnson/neural-style}}.
We used the following layers of the VGG-19 network~\cite{Simonyan14} for computing the losses: \textit{relu4\_2} for the content and \textit{relu1\_1}, \textit{relu2\_1},  \textit{relu3\_1}, \textit{relu4\_1}, \textit{relu5\_1} for the style. The energy function was minimized using L-BFGS. 
For precise evaluation we incorporated the following strict stopping criterion: the optimization was considered converged if the loss did not change by more than $0.01\%$ during 50 iterations.
This typically resulted in roughly $2000$ to $3000$ iterations for the first frame and roughly $400$ to $800$ iterations for subsequent frames when optimizing with our temporal constraint, depending on the amount of motion and the complexity of the style image.
Using a convergence threshold of $0.1\%$  cuts the number of iterations and the running time in half, and we found it still produces reasonable results in most cases. However, we used the stronger criterion in our experiments for the sake of accuracy.

For videos of resolution $350 \times 450$ we used weights $\alpha = 1$ and $\beta = 20$ for the content and style losses, respectively (default values from \emph{neural-style}),
and weight $\gamma = 200$ for the temporal losses. 
However, the weights should be adjusted if the video resolution is different.
We provide the details in section~\ref{sec:weightings}.

We ran our multi-pass algorithm with $100$ iterations per pass and set $\delta = 0.5$. Since at least $10$ passes are needed for good results, this algorithm needs in total more computation time than the single-pass approaches.

Optical flow was computed with the DeepMatching, DeepFlow, and EpicFlow implementations provided by the authors of these methods. 
We used the "improved-settings" flag in DeepMatching 1.0.1 and the default settings for DeepFlow 1.0.1 and EpicFlow 1.00.

\subsubsection{Network-based approach}

\begin{table}
	\caption{The number of frames and the training schedule for the multi-frame model.}
	\centering
	\begin{tabular}{c|c}
		Iteration & Number of frames \\
		\hline
		0 -- 60,000 & 2 \\
		60,001 -- 70,000 & 3 \\
		70,001 -- 120,000 & 5 \\
	\end{tabular}
	\label{tab:training-schedule}
\end{table}

\begin{table*}[h!]
	\centering
	\begin{tabular}{ll|lllll|c}
	    \hline\noalign{\smallskip}
		Type         & Method           & alley\_2         & ambush\_5       & ambush\_6      & bandage\_2 & market\_6 & Runtime \\ 
	    \noalign{\smallskip}\hline\noalign{\smallskip}
		Optimization & Random init      & 0.019            & 0.027  & 0.037  & 0.0180  & 0.023  & 540 s \\
		Optimization & Prev frame init  & 0.010            & 0.018  & 0.028  & 0.0041  & 0.014  & 260 s  \\
		Optimization & Ours             & \textbf{0.00061} & 0.0062 & 0.012  & \textbf{0.00084} & \textbf{0.0035} & 180 s \\
		\noalign{\smallskip}\hline\noalign{\smallskip}
		Network      & Per-frame        & 0.0062  & 0.011  & 0.016  & 0.0043  & 0.0089 & 0.2 s    \\
		Network      & Ours             & 0.0016  & \textbf{0.0042} & \textbf{0.0079} & 0.0015  & 0.0039 & 0.4 s    \\
		\noalign{\smallskip}\hline
	\end{tabular}
	\caption{Short-term temporal consistency of video style transfer. We report average mean squared error over the test set (lower is better). Pixel values were between $0$ and $1$. Run time is reported in seconds per frame. We show the results of our optimization-based and network-based approaches, as well as three baselines: optimization-based stylization initialized with random noise or the previous frame, respectively, and independent per-frame network-based processing.}
	\label{tab:video-baselines}
\end{table*}

\begin{table*}[h!]
	\centering
	\begin{tabular}{l|lllll|l}
    \hline\noalign{\smallskip}
		Subject & Nude & Picasso & Scream & Shipwreck & Woman Hat & Sum\\
		\noalign{\smallskip}\hline\noalign{\smallskip}
		Better reflects the style & 44/\textbf{55}/0 & 28/\textbf{72}/0 & 34/\textbf{61}/1 & 39/\textbf{60}/1 & 46/\textbf{53}/1 & 191/\textbf{301}/3\\
        Less flickering & 34/\textbf{53}/12 & 23/\textbf{68}/9 & 26/\textbf{59}/11 & 31/\textbf{63}/6 & 31/\textbf{62}/7 & 145/\textbf{306}/45\\
        Overall preference & 30/\textbf{68}/1 & 12/\textbf{86}/2 & 15/\textbf{80}/1 & 21/\textbf{76}/3 & 31/\textbf{66}/3 & 109/\textbf{376}/10\\
		\noalign{\smallskip}\hline
	\end{tabular}
	\caption{Results of the Amazon Mechanical Turk user study. The first number shows how many users picked the optimization-based version, the second one how many picked feed-forward and the last number represent the number of users who picked ``undecided''.}
	\label{tab:userstudy}
\end{table*}

Our network-based style transfer implementation\footnote{\url{https://github.com/manuelruder/fast-artistic-videos}} is based on Johnson et al.~\cite{Johnson2016}'s Torch implementation, called \textit{fast-neural-style}\footnote{\url{https://github.com/jcjohnson/fast-neural-style}}.
The architecture follows the design principle of Johnson et al., and we used the following layers of the VGG-16 network for the perceptual loss: \textit{relu3\_3} for the content and \textit{relu1\_2}, \textit{relu2\_2}, \textit{relu1\_2},  \textit{relu3\_3}, \textit{relu4\_3} for the style target. In the proposed style transfer network, we exchanged the upsampling convolution by two nearest neighbor upsampling layers with a regular convolution layer in between. This greatly reduces the checkerboard patterns visible in the original work. In addition, we replaced the batch normalization 
by instance normalization, as proposed in \cite{DBLP:journals/corr/UlyanovVL16}, and used a single padding layer in the beginning. The network was trained with the ADAM optimizer, a learning rate of $0.001$, a batch size of $4$ for the strided convolution architecture, and a learning rate of $0.0005$ with a batch size of $2$ for the dilated convolution variant due to memory restrictions. We trained the two-frame model for 60,000 iterations. The multi-frame model was fine-tuned from a pre-trained two-frame model. The number of iterations and frames are shown in Table \ref{tab:training-schedule}. For spherical image fine-tuning, we started from the trained multi-frame model and trained for another 40,000 iterations using the specialized dataset described in section~\ref{sec:vr-training-data}. To calculate optical flow for the videos from the Hollywood2 dataset, we used FlowNet~2.0 \cite{IMKDB17}, which is a recent state-of-the-art for optical flow estimation running at interactive frame rates.
To overcome GPU memory limitations when training with dilated convolutions, we downsampled the frames by a factor of $0.65$. We set the weight of the temporal consistency to $100$ when using mixed training and to $50$ otherwise.

\subsection{Comparison to baselines}

We benchmarked the following video stylization approaches: the optimization-based approach with per-frame random initialization, with initialization from the previous frame, and with our short-term consistency loss, where optical flow was provided by DeepFlow. Moreover, we ran the frame-wise network-based approach and our multi-frame network.

Quantitative results are shown in Table~\ref{tab:video-baselines}, averaged for $6$ style templates also used by Gatys et al. shown in Section~\ref{sec:styles-optapr}. Among optimization-based approaches, the most straightforward method~-- processing every frame independently~-- performs roughly an order of magnitude worse than the version with temporal consistency. 
Initialization with the previous stylized frame leads to better results than random initialization, but is still $2.3$ to $16$ times worse than our approach, depending on the scene.

The baseline for our network-based approach (independent per-frame) shows less flickering than the previous baseline, since this approach is more deterministic by design. Nevertheless, we consistently outperform this baseline by a factor of two or more. Interestingly, the network-based approach even outperforms the optimization-based approach in some cases while being much faster. The network-based method is more accurate on the \textit{ambush} scenes.
These include very large motion, which leads to more errors in the optical flow.
We conclude that the network-based method is able to partially correct for these errors in optical flow.

\subsection{Results of a user study}

In the previous section we have only evaluated the temporal consistency, but not the perceptual quality of the stylized videos.
We performed a comprehensive user study using Amazon Mechanical Turk (AMT) to evaluate the perceptual quality of videos stylized with our two best approaches: optimization-based with DeepFlow and network-based with multi-frame and mixed training (for details see Sections \ref{sec:detail-opt} and \ref{sec:detail-fast}).
We performed $20$ trials for each combination out of the five tested scenes and the five tested styles, resulting in a total of $500$ trials. 
In each trial a user was presented the style template, a still frame from the original video and two stylized versions of the video clip, one video processed with the optimization-based approach and the other processed with a network. 
The order of the videos was randomized. 
Inspired by~\cite{Gupta2017}, users were asked three questions: ``Which of the two videos better reflects the style of the painting?'', ``Which of the two videos has less flickering?'' and ``Overall, which of the two videos do you think looks better?''. Users had to select either one of the videos or ``undecided''. 

Table~\ref{tab:userstudy} shows the results of the user study, separately for each style image.
There was a quite clear preference towards video clips generated with the network approach, even though this approach is just an approximation of the much costlier optimization problem. This could be explained by the fact that our network approach is more robust to errors in the optical flow and can fill disocclusions more consistently due to less randomness. This reduces artifacts compared to the optimization-based approach.

\subsection{Detailed analysis of the optimization-based approach}
\label{sec:detail-opt}

\begin{table*}
	\centering
	\begin{tabular}{lll|lllll}
		\hline\noalign{\smallskip}
		initialization & penalizer & opt. flow & alley\_2 & ambush\_5 & ambush\_6 & bandage\_2 & market\_6\\
		\noalign{\smallskip}\hline\noalign{\smallskip}
		prev & no & n/a        & 0.010 & 0.018 & 0.028 & 0.0041 & 0.014 \\
		prev warped & no & DeepFlow  & 0.0016 & 0.0063 & \textbf{0.012} & 0.0015 & 0.0049 \\
		prev warped & yes & DeepFlow & \textbf{0.00061} & \textbf{0.0062} & \textbf{0.012} & 0.00084 & 0.0035 \\
		prev warped & yes & EpicFlow & 0.00073 & 0.0068 & 0.014 & \textbf{0.00080} & \textbf{0.0032} \\
		\noalign{\smallskip}\hline\noalign{\smallskip}\noalign{\smallskip}
	\end{tabular}
	\caption{Ablation study for the optimization-based approach. We report average mean squared error over the test set (lower is better). Pixel values in images were scaled between $0$ and $1$.}
	\label{tab:optimization-based}
\end{table*}

We performed an ablation study of the variants of our optimization-based approach to show the contribution of the different components and the choice of the optical flow method.
The results are provided in Table~\ref{tab:optimization-based}.
Initialization with the motion-compensated previous frame performs already much better than initializing just with the previous frame. 
The explicit temporal penalty additionally improves over this result. 
Comparing the two optical flow techniques, on average DeepFlow yields slightly better results than EpicFlow.

\begin{figure}
	\centering
	\includegraphics[width=1.0\linewidth]{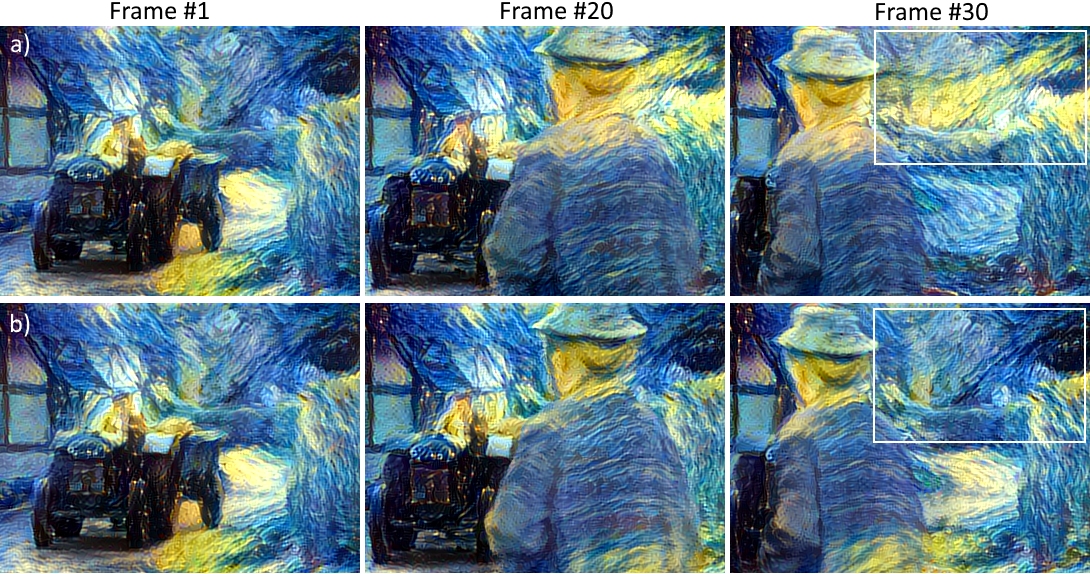}
	\caption{Scene from Miss Marple, combined with The Starry Night painting. \textbf{a)} Short-term consistency only. \textbf{b)} Long-term consistency with $J = \{1, 10, 20, 40\}$.
	The corresponding video is available at \url{https://youtu.be/SKql5wkWz8E\#t=1m40s}.}
	\label{fig:ImageLongTerm}
\end{figure}

\begin{figure}
	\centering
	\includegraphics[width=1.0\linewidth]{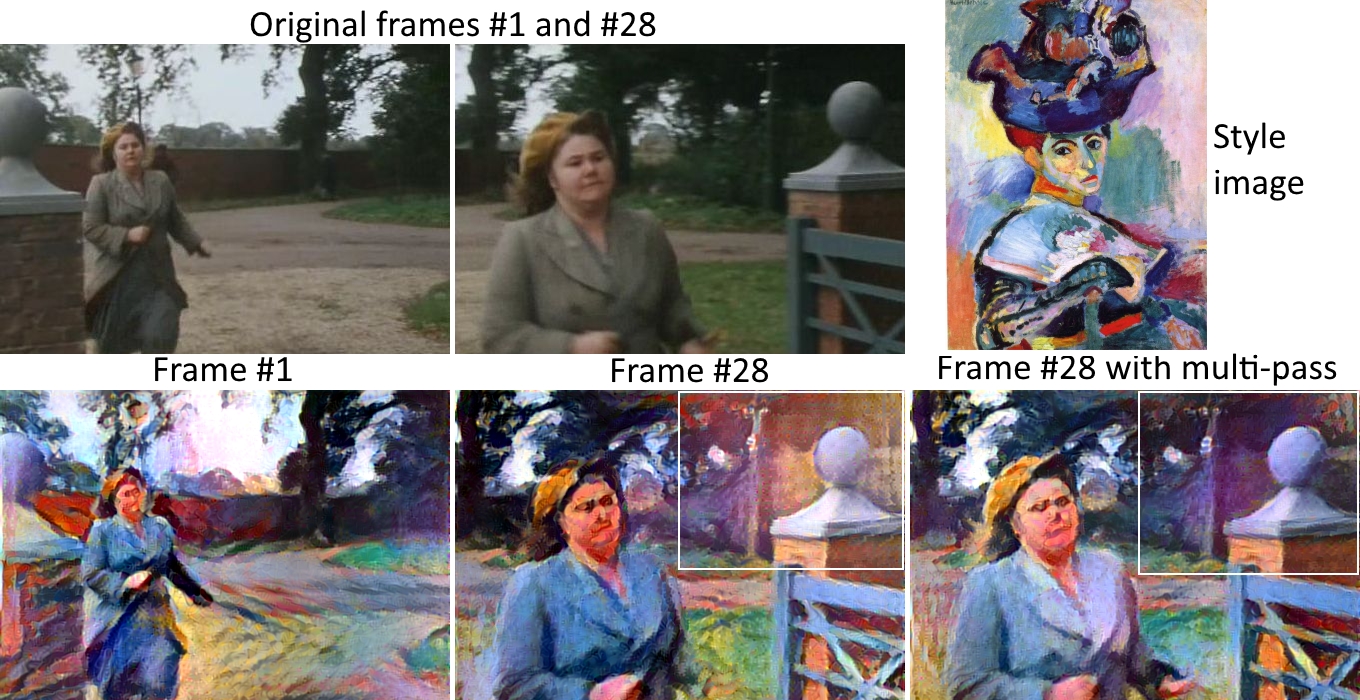}
	\caption{The multi-pass algorithm applied to a scene from Miss Marple. With the default method, the image becomes notably brighter and loses contrast, while the multi-pass algorithm better preserves the image quality over time. 
	The corresponding video can be found at \url{https://youtu.be/SKql5wkWz8E\#t=2m17s}.}
	\label{fig:ImageMultiPass}
\end{figure}

To show the benefit of the proposed long-term consistency and multi-pass technique, we cannot run a quantitative evaluation, since there is no long-term ground truth optical flow available. 
Thus, we present qualitative results in Fig.~\ref{fig:ImageLongTerm} and Fig.~\ref{fig:ImageMultiPass}, respectively. Please see also the supplementary video.

Fig.~\ref{fig:ImageLongTerm} shows a scene with a person walking through the scene. Without our long-term consistency model, the background looks very different after the person passed by. The long-term consistency model keeps the background unchanged. Fig.~\ref{fig:ImageMultiPass} shows another scene with fast camera motion. The multi-pass algorithm avoids the artifacts introduced in new image areas over time by the basic algorithm.

\subsection{Detailed analysis of the network-based approach}
\label{sec:Dilated}
\label{sec:detail-fast}

\begin{figure}
	\centering
    \includegraphics[width=1.0\linewidth]{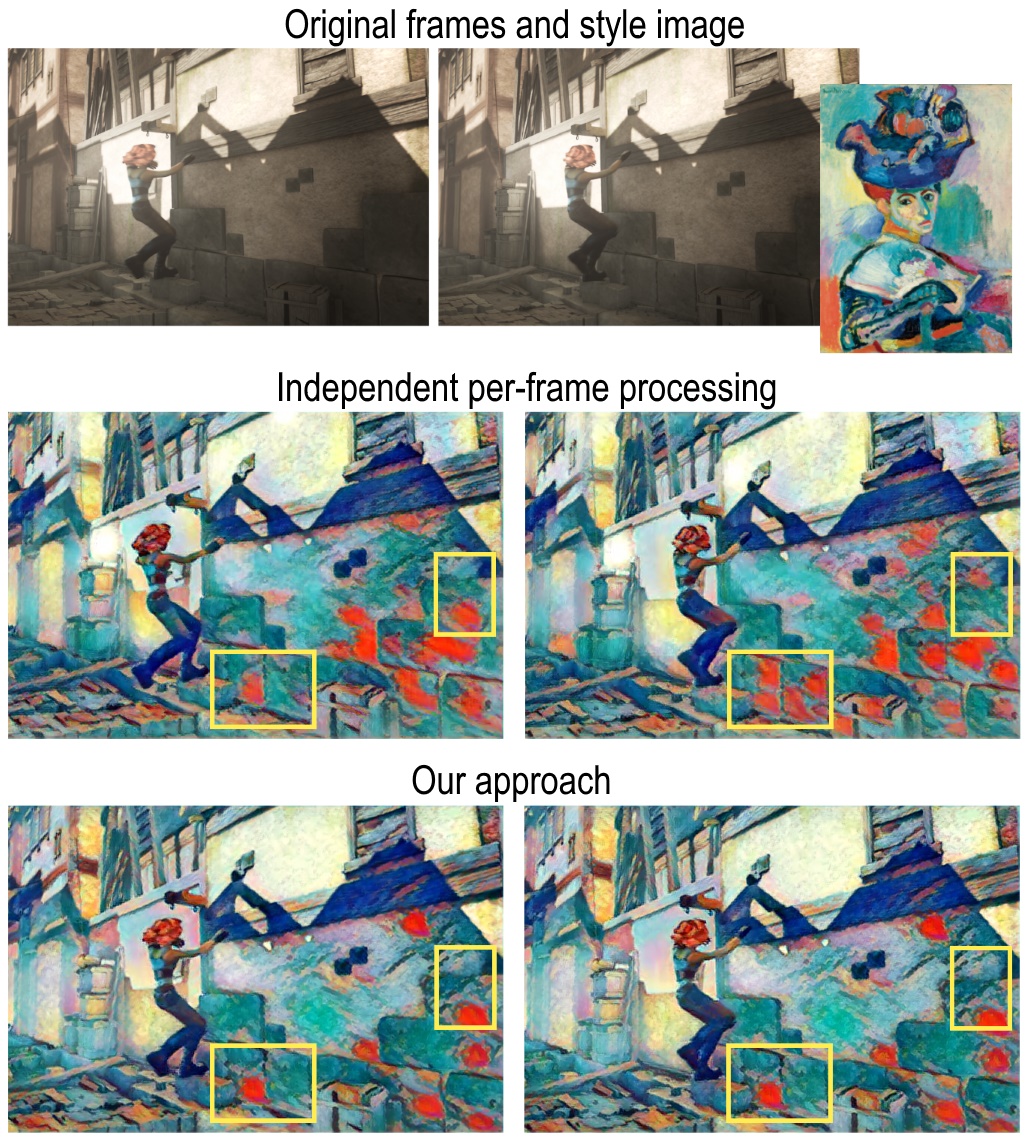}
	\caption[Comparison of temporal consistency]{Comparison of temporal consistency: Our method (multi-frame mixed) has less flickering than independent per-frame processing \cite{Johnson2016}. Rectangles indicate relevant differences. The corresponding video is available at \url{https://youtu.be/SKql5wkWz8E\#t=0m42s}}
	\label{fig:Comparison2}
\end{figure}

\begin{figure}
	\centering	
    \includegraphics[width=1.0\linewidth]{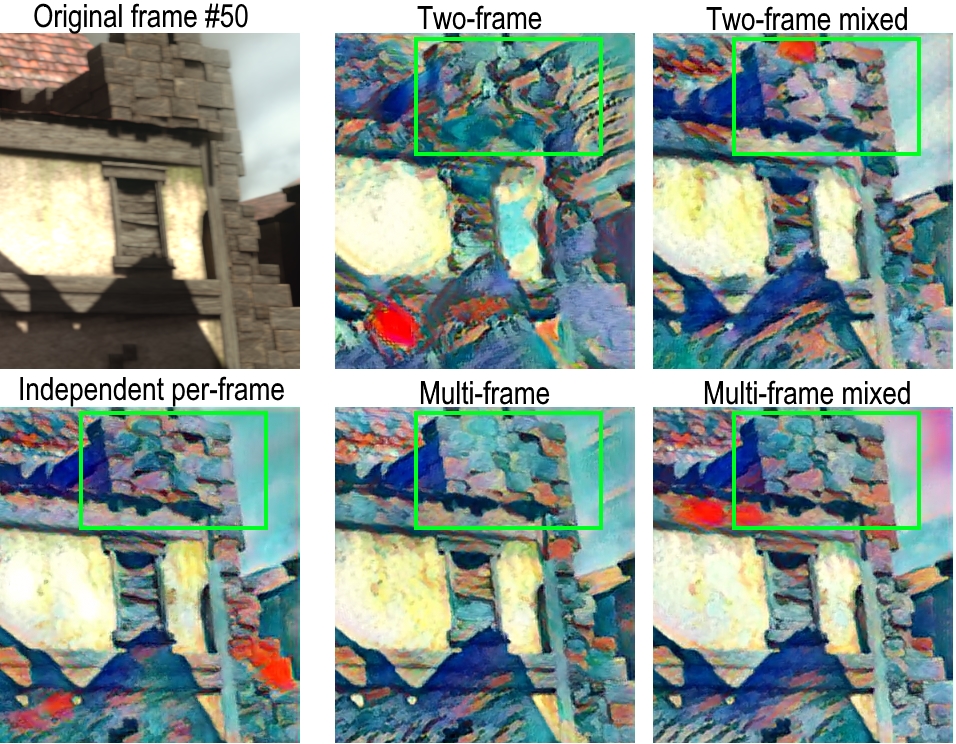}
	\caption[Comparison of different methods to counteract the propagation of error.]{Comparison of quality: Our advanced approaches retain visual quality and produce a result similar to \cite{Johnson2016} applied per-frame. The straightforward two-frame training, though, suffers from degeneration of quality. Rectangles indicate relevant differences. The corresponding video is available at \url{https://youtu.be/SKql5wkWz8E\#t=2m47s}}
	\label{fig:Comparison1}
\end{figure}

Table~\ref{tab:network-based} shows the result of an ablation study for different variants of our network-based approach. We measure temporal, style and content loss. The latter two losses aim to measure propagation of error. If the image starts to blur or degenerate over time, this should increase the style and content loss. 
The numbers show results on the complete Sintel training dataset averaged over all scenes, frames and styles. We used a different set of style templates; inspired by \cite{Johnson2016}, see Section~\ref{sec:styles-netapr}. Since the absolute losses vary heavily, we rather averaged the loss as percentage of the loss of the baseline, which is the independent per-frame processing approach.

\begin{table}
	\caption[Evaluation result on the Sintel MIP dataset.]{Comparison of network-based approaches. Values are shown as a percentage of the baseline approach \textit{BL} (independent per-frame processing). The following methods are evaluated: Per-frame processing with dilated convolutions (Di), two-frame training (2F), multi-frame training (MF) and mixed training as described in section \ref{sec:mixed-training} (2Fm, MFm).}
	\centering
	\resizebox{.5\textwidth}{!}{
    	\begin{tabular}{c|rrrrrr}
    		\hline\noalign{\smallskip}
    		\textit{Loss} & \multicolumn{1}{c}{BL} & \multicolumn{1}{c}{Di} & \multicolumn{1}{c}{2F} & \multicolumn{1}{c}{MF} & \multicolumn{1}{c}{2Fm} & \multicolumn{1}{c}{MFm} \\ 
    		\noalign{\smallskip}\hline\noalign{\smallskip}
    		Content & 100\% & \textbf{97\%} & 118\% & 101\% & 116\% & \textbf{97\%} \\ 
    		Style & 100\%  & 97\% & 167\%  & 85\% &  98\% & \textbf{77\%} \\ 
    		Temporal & 100\%  & 67\% & \textbf{41\%} & 43\% & 48\% & 49\%\\
    		\noalign{\smallskip}\hline
    	\end{tabular}
	}
	\label{tab:network-based}
	\label{tab:dilated}
\end{table}

Dilated convolution, our proposal to achieve consistency without using prior images, yields a noticeable improvement in temporal consistency while keeping the style and content losses constant, see Table~\ref{tab:dilated}. To overcome GPU memory limitations, the frames were downsampled by a factor $0.65$. To get a fair comparison, we also downsampled the frames for strided convolution and show results relative to this downsampled baseline.

The two-frame training procedure yields very good temporal consistency, but at the same time leads to a large increase in the content and the style loss. 
The best results are achieved with the mixed and multi-frame training. Both techniques are complementary and bring down the content and style loss compared to the two-frame training, while only marginally reducing temporal consistency.

Figures~\ref{fig:Comparison1} shows a qualitative comparison, another comparison can be seen in the appendix (Figure~\ref{fig:ComparisonSupp}) showing different style templates. The difference between the multi-frame and the mixed training is very subtle, while the two-frame, non-mixed training produces clearly inferior stylizations. An example on how our approach minimizes inconsistencies is shown in Figure \ref{fig:Comparison2}.

\section{Experiments on spherical images and videos}

To evaluate spherical images and videos, we downloaded 20 video scenes from YouTube containing different content (seven cityscape, six landscape and seven building interior scenes). For the spherical still image evaluation, we took three frames from each scene at a distance of approximately three seconds, i.e. 60 images.

\subsection{Spherical images}
To evaluate how well the individual cube faces fit together, i.e., if there are false discontinuities after stitching the cube faces, we measured the gradient magnitude at the cut edges (two pixel width). We report the gradient magnitude relative to the overall gradient magnitude in the image. A value of $1$ means there is no unusual gradient in that region. Values larger than $1$ indicate more gradients compared to the average gradient of the image. 

Subsequent cube faces generated with a neighboring cube face border as prior image sometimes showed false discontinuities and/or unusual high gradients at the inner edges of the prior image, where the overlap between the images end. Therefore, we also measured gradients along the inner edges of the prior image in the stylized cube faces (four pixel width). Our detailed evaluation metric is shown in the appendix.

\begin{figure}[htbp]
	\centering
	\includegraphics[width=1.0\linewidth]{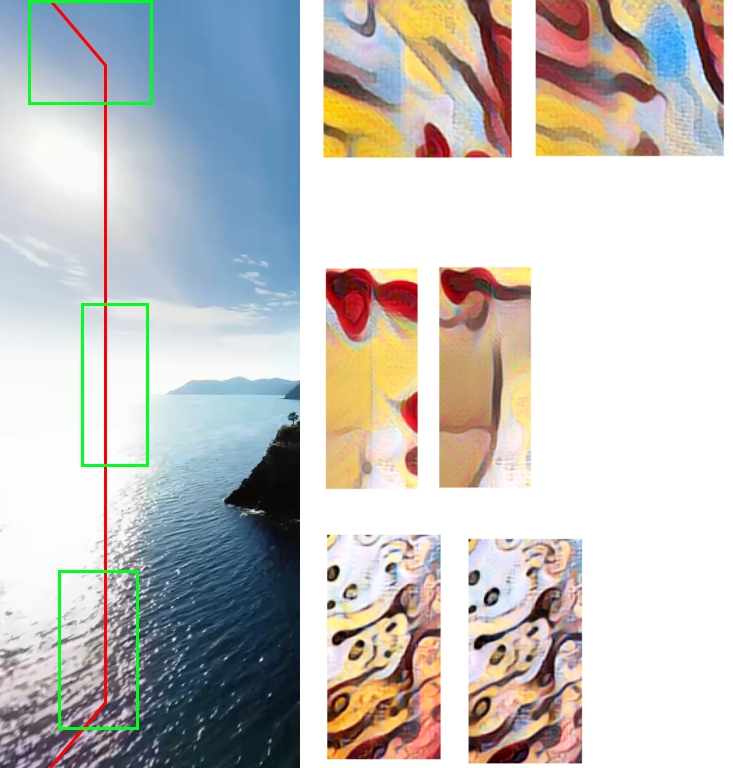}
	\caption[Closeup from an image to show how fine-tuning for spherical images can help the network to adapt to perspective transformed prior images.]{The left image shows the overlap region of a cube face from a panoramic image. The right shows close-ups for two networks. \textbf{Left:} Not fine-tuned. \textbf{Right:} Fine-tuned. In regions with little structure (top and middle), the fine-tuning strategy reduced unnatural artifacts along the inner edge of the prior image. It sometimes uses stylistic features to mask the transition (middle). In regions with more structure (bottom), both networks adapted well to the given prior.}
	\label{fig:vr-finetune-compare}
\end{figure}

Results are shown in Table~\ref{tab:vr-results}. Introducing edge consistency clearly reduces the false gradients along cut edges. 
However, it introduces new gradients along the inner edge from the prior image, most noticably in regions with little structure. 
Thus, on average, the magnitude of artificial gradients does not improve noticeably. 
Only when we use our specialized dataset for spherical images to fine-tune the model, as described in Section~\ref{sec:vr-training-data}, the gradient error improved overall in the image.
Fig.~\ref{fig:vr-finetune-compare} shows a qualitative comparison. 
In textured regions, fine-tuning is not necessary. 
In homogeneous regions, fine-tuning helps to avoid artificial edges. 
Interestingly, fine-tuning introduces style features to cover the false edges, i.e., the gradient magnitude in our quantitative measure is still increased, but the disturbing visual effect is reduced and only noticeable when large parts of the image are untextured like, for example, the top cube face showing the sky.

\begin{table}[htbp]
	\caption[Evaluation results for spherical images.]{Style transfer to spherical images. We report the gradient magnitudes at boundaries, the content and the style loss. The first column shows the baseline result when each cube face was stylized independently. "Video net" refers to a network only trained on video frames. For "Video net, ft" we fine-tuned the latter on data samples specifically made for spherical images. 
	}
	\begin{tabular}{l|ccc}
		\hline\noalign{\smallskip}
		\textit{Loss} & Ind. per face &  Video net & Video net, ft  \\ 
		\noalign{\smallskip}\hline\noalign{\smallskip}
		$E_{\text{grad}}$ & 1.46 & 1.41 & \textbf{1.21} \\
		$E_{\text{grad}_{cut}}$ & 2.48 & 1.40 & \textbf{1.17} \\
		$E_{\text{grad}_{inner}}$ & \textbf{0.95} & 1.41 & 1.23 \\
		Content & 100 \% & 101 \% & 98 \% \\
		Style & 100 \% & 104 \% & 98 \% \\
        \noalign{\smallskip}\hline
	\end{tabular}
	\label{tab:vr-results}
\end{table}

\subsection{Spherical videos}

Table~\ref{tab:video-vr-compare} shows numerical results for stylized panoramic videos at different frames of the video. We can see that unusual high gradients decrease over time, demonstrating the algorithm's ability to improve from impaired prior images. We also observe this effect when the video is mostly still and only few objects are moving.

\begin{table}[htbp]
    \caption[Evaluation results for spherical videos]{Evaluation results ($E_{\text{grad}}$ as defined above) for spherical videos. We use the multi-frame model, fine-tuned for spherical images and measure the error at different time steps.}
	\begin{tabular}{c|ccc}
		\hline\noalign{\smallskip}
		\textit{Loss} & Frame \#1 &  Frame \#2 & Frame \#10  \\ 
		\noalign{\smallskip}\hline\noalign{\smallskip}
		$E_{\text{grad}}$ & $1.21$ & $1.12$ & $0.98$ \\
		\noalign{\smallskip}\hline
	\end{tabular}
	\label{tab:video-vr-compare}
\end{table}

\subsection{Filling missing regions}

To further reduce unusual gradients, we experimented with different ways to fill in missing or disoccluded regions in the prior image. Originally, we masked missing regions with zeros, which is equivalent to the mean pixel of the ImageNet dataset\footnote{In the pretrained VGG models, which we use for our perceptual loss, values are centered around the mean pixel of the ImageNet dataset.}. By filling missing regions with random noise instead, we want to avoid that the network falsely observes sharp edges along missing regions.

Numerical analysis shows that the $E_{gradient}$ measure indeed improves slightly from $1.21$ to $1.18$ when filling masked regions with random noise.\\

\begin{figure}[htbp]
	\centering
	\includegraphics[width=0.8\linewidth]{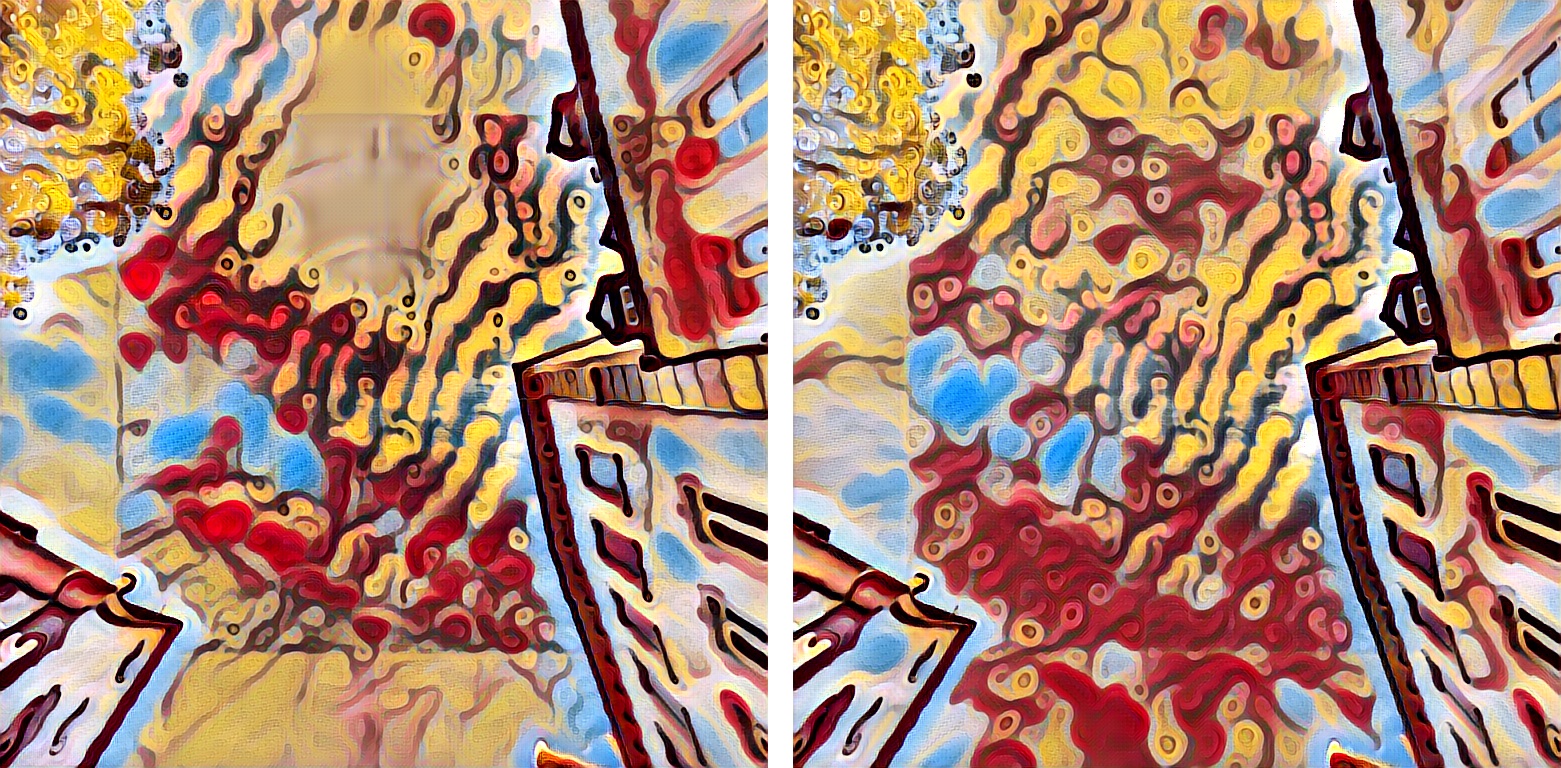}
	\caption[Comparison of the top cube face with different ways to fill masked regions]{Comparison of the top cube face with different ways to fill masked regions. Left: Zeros, right: Random noise.}
	\label{fig:VR-FillIn}
\end{figure}

Most noticeably, a random fill-in improves a failure case of our algorithm: regions with little structure in the image, see Figure~\ref{fig:VR-FillIn}. False gradients along the prior image boundary are less visible.

\begin{figure}[htbp]
	\centering
	\includegraphics[width=1.0\linewidth]{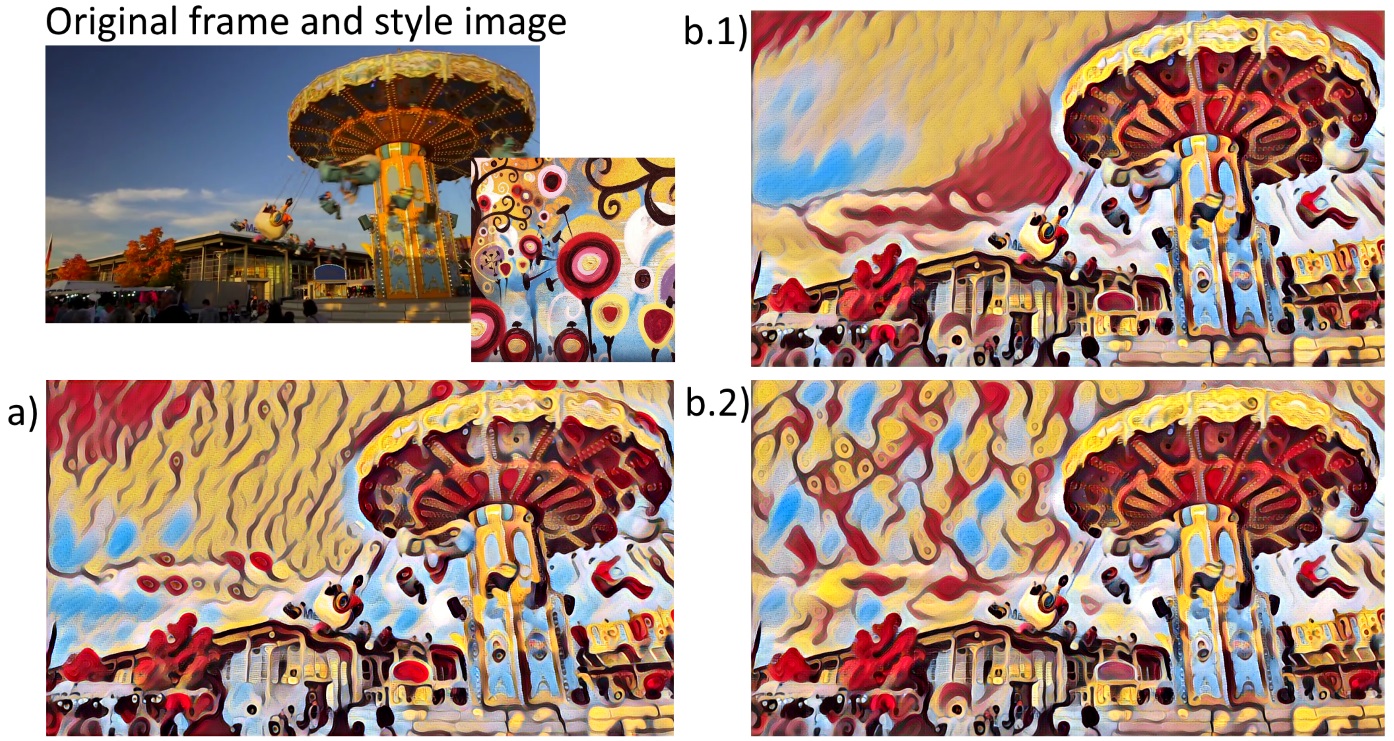}
	\caption[Effect of the occlusion fill-in.]{On the effect of different ways to fill disoccluded regions. a) Trained and inferred with zeros as fill-in. b.1) Trained with random noise, but inferred with zeros. b.2) Trained and inferred with random noise.
	Images were generated with no prior image given, but either a zero-vector or random noise, as if the whole image was disoccluded.}
	\label{fig:occFill}
\end{figure}

However, we observed that a random fill-in further changes the way the network fills missing regions. To get a better understanding of how the fill-in influences the network's behavior, we performed an experiment where we used random noise as fill-in during training, but used zeros at test time. Figure~\ref{fig:occFill} shows how this changes the output. For this experiment, we gave the network an empty prior image, as if the whole image was missing or disoccluded, i.e. the prior image is either a vector containing only zeros or random noise. Once we removed this extra source of entropy, the network was able to rely on it during training, and the image loses variance especially in regions with little structure. We conclude that the network has learned to make use of the noise such that it creates a more diverse appearance in regions with little structure.

\section{Runtimes}

We measured the runtime of our algorithms on an Nvidia Titan X GPU (Maxwell architecture) and a resolution of $1024 \times 436$ pixel.

Using the relaxed convergence threshold of $0.1\%$, the optimization-based approach takes around eight to ten minutes for the initial frame initialized with random noise. When initialized with the warped previous frame and combined with our temporal loss, the optimization converges 2 to 3 times faster, three minutes on average.

Our feed-forward approach only takes $400$ ms to stylize a new frame, including the time needed to pre-process the last frame (masking and warping, which is done on the GPU as well), but excluding time for I/O operations (e.g. writing the file to disk). This corresponds to a speedup factor of $450$ compared to the optimization-based approach. A single-image processed with Johnson et al. \cite{Johnson2016} needed $200ms$ at that resolution. Training the multi-frame model takes around 24 hours for 120,000 iterations.

For panoramic images, our approach needs $3.9$ seconds to stylize a full frame ($650ms$ for each cube face with resolution $768 \times 768$ including overlapping regions. Without overlapping, this equals a resolution of $640 \times 640$ per cube face). If the video is not distributed in the cube face projection, additional time for reprojection has to be taken into account.

\section{Conclusion}
We presented a set of techniques for optimization-based style transfer in videos: suitable initialization and a loss function that enforces short-term temporal consistency of the stylized video, a loss function for long-term consistency, and a multi-pass approach. As a consequence, we can produce stable and visually appealing stylized videos even in the presence of fast motion and strong occlusion.

Furthermore, we presented an algorithm to train a neural network with a prior image in order to create coherent style transfer with a much lower run time per frame. We were able to prevent the propagation of errors when applying our model recursively: by iterating the network multiple times during training, using a longer sequence of frames, we trained our network to recognize and counteract progressed degeneration. We also generalized our model for spherical images: in this task, the prior image is not the last frame, but an edge from a neighboring cube face, perspectively transformed by $90^{\circ}$. We could show that our model is able to adapt to such a prior image when there is enough structure in the image.

\bibliographystyle{spmpsci}
\bibliography{references}

\begin{thebibliography}{10}
\providecommand{\url}[1]{{#1}}
\providecommand{\urlprefix}{URL }
\expandafter\ifx\csname urlstyle\endcsname\relax
  \providecommand{\doi}[1]{DOI~\discretionary{}{}{}#1}\else
  \providecommand{\doi}{DOI~\discretionary{}{}{}\begingroup
  \urlstyle{rm}\Url}\fi

\bibitem{Butler:ECCV:2012}
Butler, D.J., Wulff, J., Stanley, G.B., Black, M.J.: A naturalistic open source
  movie for optical flow evaluation.
\newblock In: ECCV, pp. 611--625 (2012)

\bibitem{DBLP:journals/corr/ChenLYYH17}
Chen, D., Liao, J., Yuan, L., Yu, N., Hua, G.: Coherent online video style
  transfer.
\newblock CoRR \textbf{abs/1703.09211} (2017)

\bibitem{Collobert_NIPSWORKSHOP_2011}
Collobert, R., Kavukcuoglu, K., Farabet, C.: Torch7: A matlab-like environment
  for machine learning.
\newblock In: BigLearn, NIPS Workshop (2011)

\bibitem{DBLP:conf/nips/GatysEB15}
Gatys, L.A., Ecker, A.S., Bethge, M.: Texture synthesis using convolutional
  neural networks.
\newblock In: NIPS, pp. 262--270 (2015)

\bibitem{Gatys2016}
Gatys, L.A., Ecker, A.S., Bethge, M.: Image style transfer using convolutional
  neural networks.
\newblock In: CVPR, pp. 2414--2423 (2016)

\bibitem{Ghiasi2017}
Ghiasi, G., Lee, H., Kudlur, M., Dumoulin, V., Shlens, J.: Exploring the
  structure of a real-time, arbitrary neural artistic stylization network.
\newblock CoRR \textbf{abs/1705.06830} (2017)

\bibitem{Gupta2017}
Gupta, A., Johnson, J., Alahi, A., Fei-Fei, L.: Characterizing and improving
  stability in neural style transfer.
\newblock CoRR \textbf{1705.02092} (2017)

\bibitem{Hays:2004:IVB:987657.987676}
Hays, J., Essa, I.: Image and video based painterly animation.
\newblock In: Proceedings of the 3rd International Symposium on
  Non-photorealistic Animation and Rendering, NPAR, pp. 113--120 (2004)

\bibitem{Huang17}
Huang, H., Wang, H., Luo, W., Ma, L., Jiang, W., Zhu, X., Li, Z., Liu, W.:
  Real-time neural style transfer for videos.
\newblock In: CVPR (2017)

\bibitem{IMKDB17}
Ilg, E., Mayer, N., Saikia, T., Keuper, M., Dosovitskiy, A., Brox, T.: Flownet
  2.0: Evolution of optical flow estimation with deep networks.
\newblock In: CVPR (2017)

\bibitem{Ioffe2015BatchNA}
Ioffe, S., Szegedy, C.: Batch normalization: Accelerating deep network training
  by reducing internal covariate shift.
\newblock In: ICML (2015)

\bibitem{Johnson2016}
Johnson, J., Alahi, A., Fei{-}Fei, L.: Perceptual losses for real-time style
  transfer and super-resolution.
\newblock In: ECCV, pp. 694--711 (2016)

\bibitem{DBLP:conf/cvpr/LiW16}
Li, C., Wand, M.: Combining markov random fields and convolutional neural
  networks for image synthesis.
\newblock In: CVPR, pp. 2479--2486 (2016)

\bibitem{DBLP:conf/eccv/LiW16}
Li, C., Wand, M.: Precomputed real-time texture synthesis with markovian
  generative adversarial networks.
\newblock In: ECCV, pp. 702--716 (2016)

\bibitem{coco04}
Lin, T.Y., Maire, M., Belongie, S., Hays, J., Perona, P., Ramanan, D., Dollár,
  P., Zitnick, C.L.: Microsoft coco: Common objects in context.
\newblock In: ECCV (2014)

\bibitem{Litwinowicz:1997:PIV:258734.258893}
Litwinowicz, P.: Processing images and video for an impressionist effect.
\newblock In: Proceedings of the 24th Annual Conference on Computer Graphics
  and Interactive Techniques, SIGGRAPH, pp. 407--414 (1997)

\bibitem{luan2017deep}
Luan, F., Paris, S., Shechtman, E., Bala, K.: Deep photo style transfer.
\newblock arXiv preprint arXiv:1703.07511  (2017)

\bibitem{DBLP:conf/cvpr/MarszalekLS09}
Marszalek, M., Laptev, I., Schmid, C.: Actions in context.
\newblock In: CVPR, pp. 2929--2936 (2009)

\bibitem{DBLP:journals/corr/NikulinN16}
Nikulin, Y., Novak, R.: Exploring the neural algorithm of artistic style.
\newblock CoRR \textbf{abs/1602.07188} (2016)

\bibitem{odonovan:2012}
O'Donovan, P., Hertzmann, A.: Anipaint: Interactive painterly animation from
  video.
\newblock Transactions on Visualization and Computer Graphics \textbf{18}(3),
  475--487 (2012)

\bibitem{revaud:hal-01142656}
Revaud, J., Weinzaepfel, P., Harchaoui, Z., Schmid, C.: Epicflow:
  Edge-preserving interpolation of correspondences for optical flow.
\newblock In: CVPR, pp. 1164--1172 (2015)

\bibitem{Ruder2016}
Ruder, M., Dosovitskiy, A., Brox, T.: Artistic style transfer for videos.
\newblock In: GCPR, pp. 26--36 (2016)

\bibitem{Simonyan14}
Simonyan, K., Zisserman, A.: Very deep convolutional networks for large-scale
  image recognition.
\newblock ICLR  (2015)

\bibitem{Bro10e}
Sundaram, N., Brox, T., Keutzer, K.: Dense point trajectories by
  gpu-accelerated large displacement optical flow.
\newblock In: ECCV, pp. 438--451 (2010)

\bibitem{DBLP:conf/icml/UlyanovLVL16}
Ulyanov, D., Lebedev, V., Vedaldi, A., Lempitsky, V.S.: Texture networks:
  Feed-forward synthesis of textures and stylized images.
\newblock In: ICML, pp. 1349--1357 (2016)

\bibitem{DBLP:journals/corr/UlyanovVL16}
Ulyanov, D., Vedaldi, A., Lempitsky, V.S.: Instance normalization: The missing
  ingredient for fast stylization.
\newblock CoRR \textbf{abs/1607.08022} (2016)

\bibitem{weinzaepfel:hal-00873592}
Weinzaepfel, P., Revaud, J., Harchaoui, Z., Schmid, C.: {DeepFlow: Large
  displacement optical flow with deep matching}.
\newblock In: ICCV, pp. 1385--1392 (2013)

\bibitem{YuKoltun2016}
Yu, F., Koltun, V.: Multi-scale context aggregation by dilated convolutions.
\newblock In: ICLR (2016)

\bibitem{DBLP:journals/corr/ZhangD17}
Zhang, H., Dana, K.J.: Multi-style generative network for real-time transfer.
\newblock CoRR \textbf{abs/1703.06953} (2017)

\end{thebibliography}

\newpage
\newpage\phantom{blabla}

\appendix
\renewcommand\thefigure{\thesection.\arabic{figure}}
\setcounter{figure}{0}

\section{Appendix}

\subsection{Supplementary videos}
\label{sec:video}
A supplementary video, available at \url{https://youtu.be/SKql5wkWz8E}, shows moving sequences corresponding to figures from this paper, plus a number of additional results:
\begin{itemize}
\item Results of the optimization-based algorithm on different sequences, including a comparison of the basic algorithm and the multi-pass and long-term algorithm
\item Comparison of "naive" ($\cert$) and "advanced" ($\certlong$) weighting schemes for long-term consistency
\item Results of the feed-forward algorithm on different sequences, including results from different techniques to reduce the propagation of errors.
\item Comparison between optimization-based and fast style transfer
\item A demonstration of our panoramic video algorithm
\end{itemize}

Another video, showing a full panoramic video in $360\degree$, can be found at \url{https://youtu.be/pkgMUfNeUCQ}.

\subsection{Style images and parameter configuration}\label{sec:styles}\label{sec:weightings}

\subsubsection{Optimization-based approach}\label{sec:styles-optapr}

\begin{figure}[!ht]
	\centering
	\includegraphics[width=0.9\linewidth]{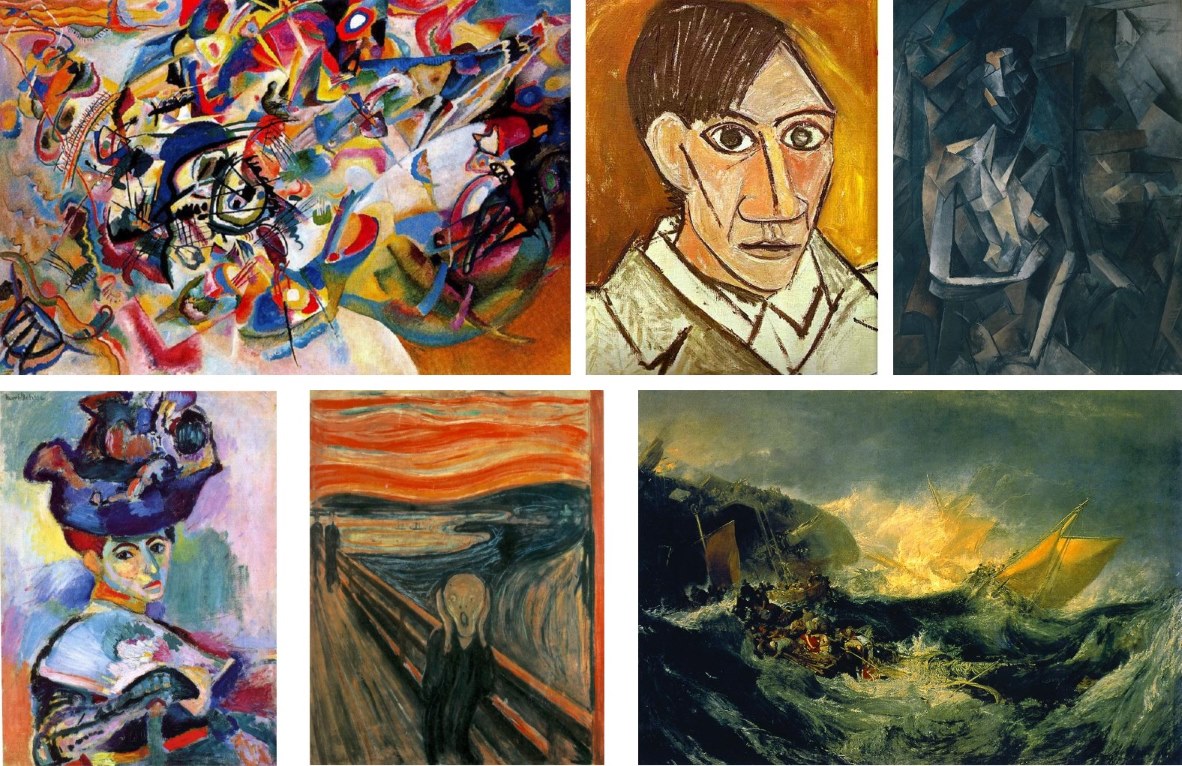}
	\caption{Styles used for experiments on Sintel. Left to right, top to bottom: "Composition VII" by Wassily Kandinsky (1913), Self-Portrait by Pablo Picasso (1907), "Seated female nude" by Pablo Picasso (1910), "Woman with a Hat" by Henri Matisse (1905), "The Scream" by Edvard Munch (1893), "Shipwreck" by William Turner (1805).}
	\label{fig:styles_benchmark}
\end{figure}

Figure~\ref{fig:styles_benchmark} shows the style images chosen to evaluate the optimization-based approach and to perform the user study (except Composition VII), inspired by the selection of styles by \cite{Gatys2016}.

\subsubsection{Network-based approach}\label{sec:styles-netapr}

\begin{figure}[!ht]
	\centering
	\includegraphics[width=0.9\linewidth]{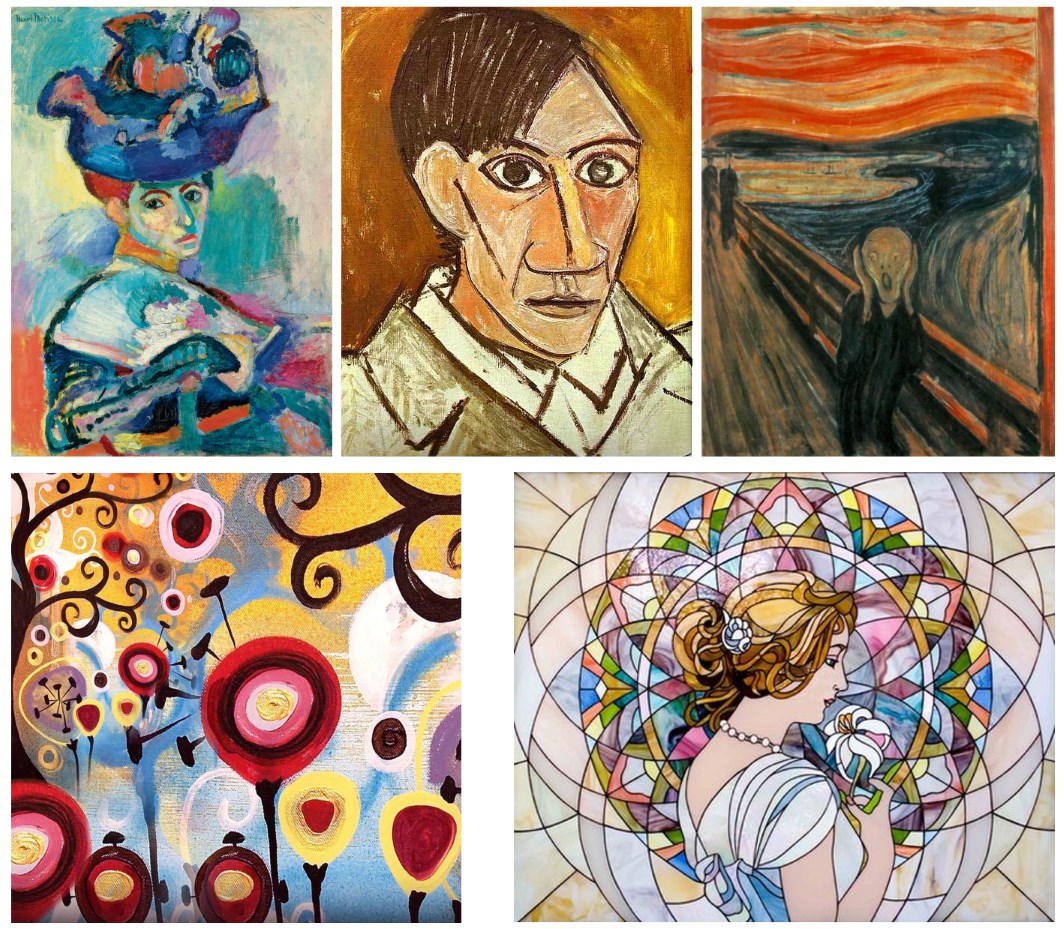}
	\caption[Style images used for the evaluation.]{Style images used for the evaluation. From left to right, top to bottom: \textit{Woman with a Hat} by Henri Matisse (1905), Self-Portrait by Pablo \textit{Picasso} (1907), \textit{The Scream} by Edvard Munch (1893), collage of the painting June Tree by Natasha Wescoat (also referred to as \textit{Candy} by Johnson et al.), glass painting referred to as \textit{Mosaic} by Johnson et al.}
	\label{fig:styles}
\end{figure}

Figure~\ref{fig:styles} shows the style images used for the detailed analysis of the network-based approach and for spherical image and video evaluation, inspired by the selection of styles by \cite{Johnson2016}. In Table~\ref{tab:style-parameters}, the parameters for training individual models are shown.

\begin{table}[!ht]
	\begin{tabular}{l|ccccc}
		\hline\noalign{\smallskip}
		Param & WomanHat & Picasso & Candy & Mosaic & Scream  \\ 
		\noalign{\smallskip}\hline\noalign{\smallskip}
		\begin{tabular}{@{}l@{}}Style \\ Scale\end{tabular} & 384 & 384  & 384  & 512 & 384  \\ 
		\begin{tabular}{@{}l@{}}Content \\ Weight\end{tabular} & 1 & 1  & 1 & 1  & 1 \\ 
		\begin{tabular}{@{}l@{}}Style \\ Weight\end{tabular} & 20 & 10  & 10 & 10 & 20 \\
		 \noalign{\smallskip}\hline
	\end{tabular}
	\caption{The individual training parameters for the style images.}
	\label{tab:style-parameters}
\end{table}

\subsection{Effect of errors in optical flow estimation}

\begin{figure}[h]
	\centering
	\includegraphics[width=1\linewidth]{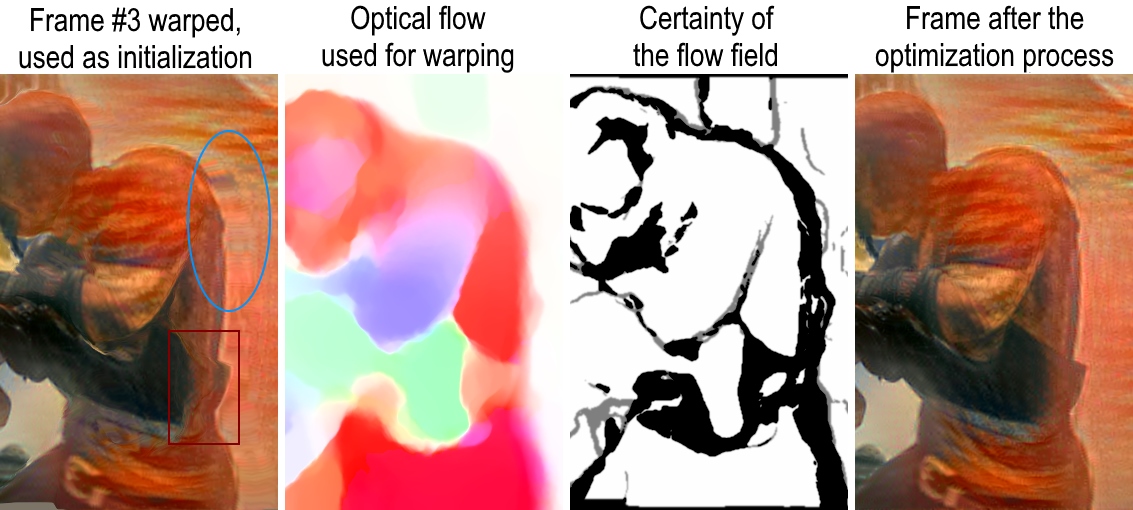}
	\caption{Scene from the Sintel video showing how the algorithm deals with optical flow errors (red rectangle) and disocclusions (blue circle). Both artifacts are somehow repaired in the optimization process due to the exclusion of uncertain areas from our temporal constraint.
		Still, optical flow errors lead to imperfect results. The third image shows the uncertainty of the flow filed in black and motion boundaries in gray.}
	\label{fig:Image2}
\end{figure}

The quality of results produced by our algorithm strongly depends on the quality of optical flow estimation.
This is illustrated in Figure~\ref{fig:Image2}. 
When the optical flow is correct (top right region of the image), the method manages to repair the artifacts introduced by warping in the disoccluded region.
However, erroneous optical flow (tip of the sword in the bottom right) leads to degraded performance. The optimization process partially compensates for the errors (sword edges get sharp), but cannot fully recover.

\subsection{Robust loss function for temporal consistency in the optimization-based approach}\label{sec:robust_loss}

\begin{figure}
\centering
\includegraphics[width=0.9\linewidth]{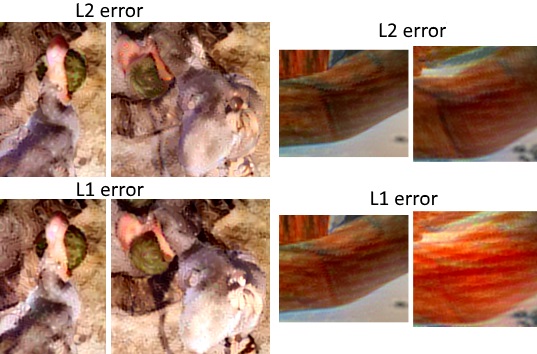}
\caption{\textbf{Left}: Scene from Ice Age (2002) where an absolute error function works better, because the movement of the bird was not captured correctly by the optical flow. \textbf{Right}: Extreme case from Sintel movie where a squared error is clearly superior.}
\label{fig:Image3}
\end{figure}

We tried using the more robust absolute error instead of squared error for the temporal consistency loss. The weight for the temporal consistency was doubled in this case.
Results are shown in Figure~\ref{fig:Image3}.
While in some cases (left example in the figure) the absolute error leads to slightly improved results, in other cases (right example in the figure) it causes large fluctuations.
We therefore stick with the mean squared error in all our experiments.

\subsection{Batch and instance normalization.}
Batch normalization \cite{Ioffe2015BatchNA} normalizes the feature activations for individual feature maps in a mini-batch after each layer of a neural network. This has been found to be advantageous especially for very deep neural networks, where the variance over the output is likely to shift during training.
Let $x \in \mathbb{R}^{B \times C \times H \times W}$ be a tensor with batch size $B$, $C$ channels and spatial dimensions $H \times W$, and let $x_{bchw}$ be the $bchw$-th element in this tensor. Then, the mean and the variance are calculated as $\mu_{c} = \frac{1}{BHW} \sum_{b=1}^{B} \sum_{h=1}^{H} \sum_{w=1}^{W} x_{bchw}$ and
$\sigma_{c}^2 = \frac{1}{BHW} \sum_{b=1}^{B} \sum_{h=1}^{H} \sum_{w=1}^{W} (x_{bchw} - \mu_{c})^2$.

The batch normalization layer performs the following operation to compute the output tensor $y$:
\begin{equation} \label{eq:BN-2}
y_{bchw} = \gamma \frac{x_{bchw} - \mu_{c}}{\sqrt{\sigma_{c}^2 + \epsilon}} + \beta,
\end{equation}
with learnable scale and shift parameters $\gamma$ and $\beta$.

In contrast, instance normalization normalizes every single instance in a batch, that is, contrast normalization is performed for each individual instance. Therefore, the mean and the variance are computed per instance and feature map. We define a separate mean and variance for each instance as $\mu_{bc} = \frac{1}{HW} \sum_{h=1}^{H}\sum_{w=1}^{W} x_{bchw}$ and $\sigma_{bc}^2 = \frac{1}{HW}\sum_{h=1}^{H}\sum_{w=1}^{W} (x_{bchw} - \mu_{bc})^2$.

The instance normalization layer performs the following operation to compute the output tensor $y$:
\begin{equation}
y_{bchw} = \frac{x_{bchw} - \mu_{bc}}{\sqrt{\sigma_{bc}^2 + \epsilon}}.
\end{equation}

\subsection{Comparison of different methods to reduce the propagation of error in the network approach}

\begin{figure}[htbp]
	\centering	
    \includegraphics[width=1.0\linewidth]{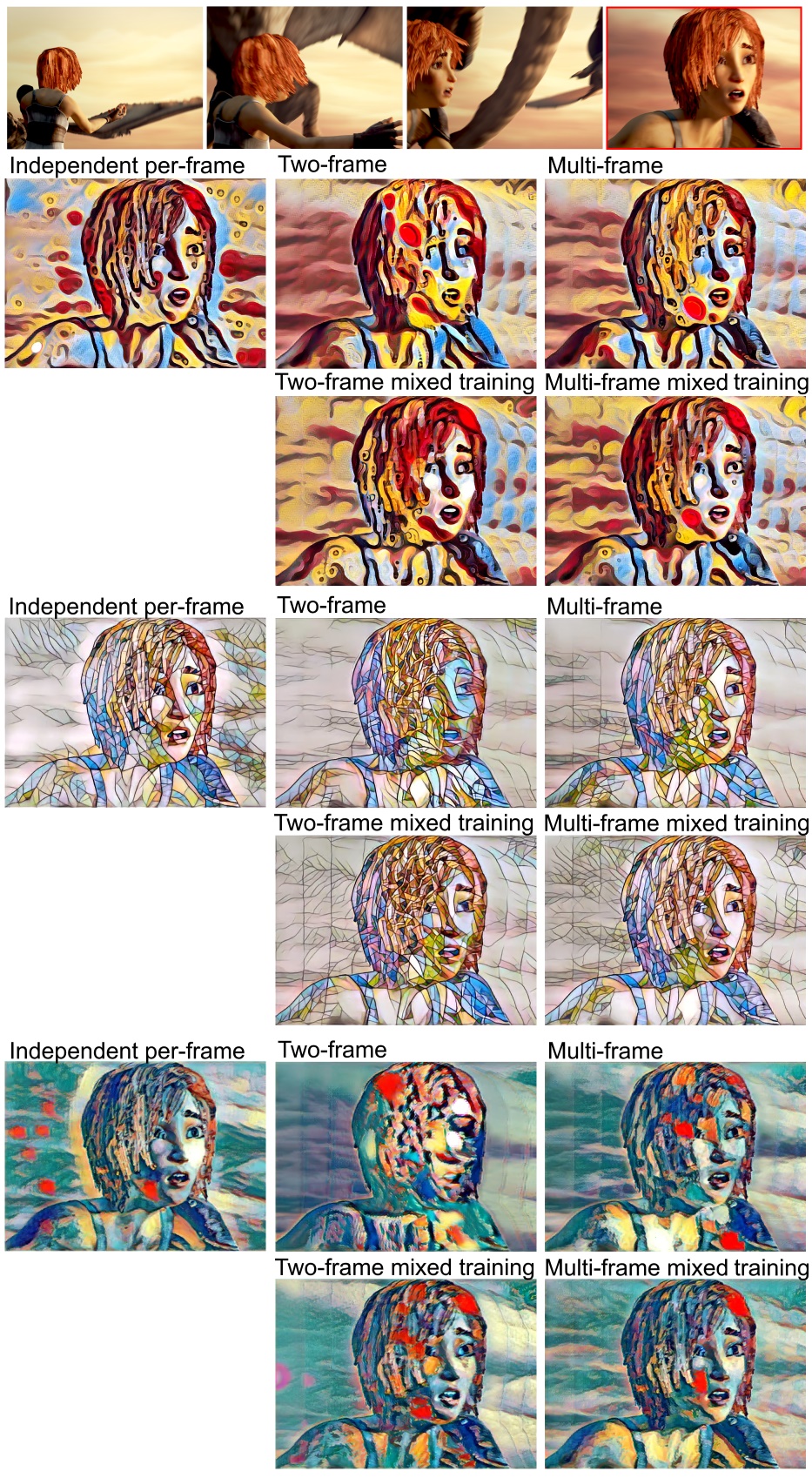}
	\caption[Comparison of different methods to counteract the propagation of error.]{Comparison of quality: Even for scenes with fast-motion, our advanced approaches retain visual quality and produce a result similar to Johnson et al. applied per-frame. The straightforward two-frame training, though, suffers from degeneration of quality.}
	\label{fig:ComparisonSupp}
\end{figure}

\subsection{Convergence of our style transfer network}

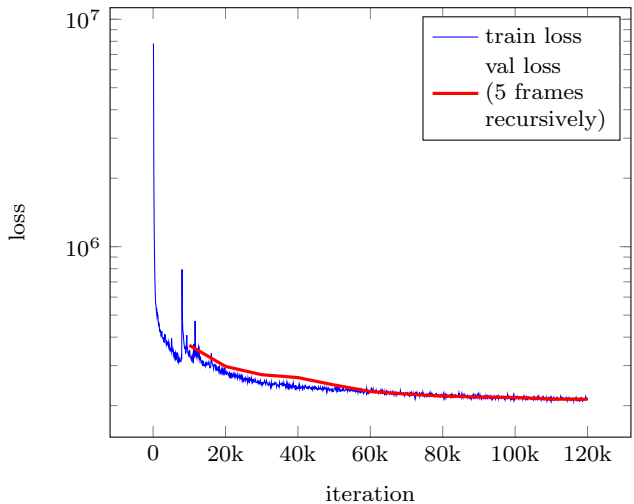
\begin{figure}[!ht]
	\centering
    \begin{tikzpicture}
    \begin{axis}[ymode=log, log basis y={10}, enlargelimits=true, xlabel=iteration, ylabel=loss, scaled ticks=false, tick label style={/pgf/number format/fixed}, xtick = {0,20000,40000,60000,80000,100000,120000},xticklabels ={0,20k,40k,60k,80k,100k,120k},legend style={cells={align=left}},legend cell align=left]
    \addplot[blue, mark size=0.5] table [x=iteration, y=loss, col sep=comma]  {data-train.csv};
    \addlegendentry{train loss}
    \addplot[red, mark size=1.5, line width=1.2pt] table [x=iteration, y=loss, col sep=comma, legend pos=north east]  {data-val.csv};
    \addlegendentry{val loss\\($5$ frames\\recursively)}
    \end{axis}
    \end{tikzpicture}
	\caption[Training and validation loss]{Training loss (blue) and validation loss (red) for a multi-frame training on a logarithmic scale. Length of the frame sequence used for training, depending on the iteration number: 0-50k: $2$; 50k-60k: $3$; 60k-120k: $5$. The validation loss is computed every 10k iterations (starting from iteration 10k) and is always the average loss for $5$ consecutive frames processed recursively. Therefore the validation loss is larger than the training loss in the beginning, but decreases as our multi-frame training begins.}
	\label{fig:convergence}
\end{figure}

Our network converges without overfitting, as seen in Figure~\ref{fig:convergence}. For the validation loss, we always use the average loss of $5$ consecutive frames, processed recursively, so that the validation objective stays constant during the training of the multi-frame model. We used 120,000 iterations for the sake of accuracy, but Figure~\ref{fig:convergence} also indicates that the training can be stopped earlier in practice.

\subsection{Reprojection for border consistency in spherical images}
For perspective transformation of a border region we virtually organize the cube faces in a three dimensional space, so that we can use 3D projection techniques such as the pinhole camera model. According to the pinhole camera model, for a point $(x_1, x_2, x_3)$ in a three-dimensional Cartesian space, the projection $(y_1, y_2)$ on the target plane is calculated as:
\begin{equation}
{{y_1}\choose{y_2}} = -\frac{d}{x_3} {{x_1}\choose{x_2}},
\end{equation}
where $d$ is the distance from the projection to the origin. We arrange the already stylized cube face at an angle of $90\degree$ to the projection plane to project its border into the plane of another cube face.

\subsection{Evaluation metric for spherical images}

To evaluate if there are unusually high gradients in a given region compared to the rest of the image, we calculate the ratio of gradient magnitudes in that region to gradient magnitudes in the overal image.
We take the maximum color gradients per pixel, i.e. we define our image gradient in $x$ direction as $G_x(\vec{p}) = \operatorname{cmax}\Bigl(\vec{p}[red]_x, \vec{p}[green]_x, \vec{p}[blue]_x\Bigr)$. On that basis, we define the ratio $r_x =\dfrac{|| \ \vec{s}_{x} \circ G_{x} ||_1}{|| \vec{s}_{x} ||_1}\Big/ \dfrac{|| G_{x} ||_1}{D}$ where $D$ is the dimensionality of the image and $\vec{s}$ is a binary vector which encodes the region where we want to test for unusual gradients (in $x$ direction), such that the vector is $1$ in the desired region and $0$ everywhere else. By $\circ$ we denote the element-wise multiplication.

The error metric $E_{gradient}$, which is a weighted average between horizontal and vertical gradients, is then calculated as follows:
\begin{equation}
E_{gradient} =  \frac{ ||\vec{s}_{x}||_1 \cdot r_x + ||\vec{s}_{y}||_1 \cdot r_y  }{||\vec{s}_{x}||_1 + ||\vec{s}_{y}||_1}
\end{equation}

\end{document}